%% file: main.tex
\documentclass[11pt, oneside]{article}

\input{preamble.tex}

\usepackage{credits}
\usepackage{orcidlink}
\usepackage{hyperref}
\usepackage{multirow}
\usepackage{threeparttable}
\usepackage{arydshln}
\usepackage{xcolor}
\usepackage{booktabs}
\usepackage{tcolorbox}

\title{\textsf{Probing the limitations of multimodal language models for chemistry and materials research}}

\input{authors.tex}

\begin{document}
\maketitle
 
\begin{abstract}
\noindent Recent advancements in artificial intelligence have sparked interest in scientific assistants that could support researchers across the full spectrum of scientific workflows, from literature review to experimental design and data analysis. 
A key capability for such systems is the ability to process and reason about scientific information in both visual and textual forms---from interpreting spectroscopic data to understanding laboratory setups. Here, we introduce MaCBench, a comprehensive benchmark for evaluating how vision-language models handle real-world chemistry and materials science tasks across three core aspects: data extraction, experimental understanding, and results interpretation. 
Through a systematic evaluation of leading models, we find that while these systems show promising capabilities in basic perception tasks—achieving near-perfect performance in equipment identification and standardized data extraction---they exhibit fundamental limitations in spatial reasoning, cross-modal information synthesis, and multi-step logical inference. 
Our insights have important implications beyond chemistry and materials science, suggesting that developing reliable multimodal AI scientific assistants may require advances in curating suitable training data and approaches to training those models.
\end{abstract}

\pagebreak

\begin{refsection}
\input{sections/introduction}
\input{sections/results}
\input{sections/conclusions}
\input{sections/methods}

\input{sections/data_availability}

\input{sections/acknowledgment}

\input{sections/competing_interests}

\section*{Author contributions}
\footnotesize
\insertcredits
\normalsize 

\printbibliography
\end{refsection}
\clearpage

\renewcommand\thefigure{\thesection.\arabic{figure}}

\renewcommand\thetable{\thesection.\arabic{table}}
\renewcommand\theequation{\thesection\arabic{equation}}
\setcounter{figure}{0}
\setcounter{table}{0}
\setcounter{equation}{0}

\begin{refsection}
\appendix
\input{appendix}

\clearpage
\printbibliography
\end{refsection}

\clearpage
\printnoidxglossary[type=\acronymtype]
\printnoidxglossary[sort=letter]

\end{document}

%% file: preamble.tex
\usepackage[english]{babel}
\usepackage{geometry}  
\usepackage{graphicx}      
\usepackage{tabularx}      
\usepackage{amssymb}
\usepackage{amsmath}
\usepackage{eulervm}
\usepackage{xltabular}
\usepackage[hyphens]{url}
\usepackage[hidelinks]{hyperref}
\hypersetup{breaklinks=true}
\usepackage[backend=biber,style=nature, doi=false,isbn=false, date=year, url=false, autopunct=true]{biblatex}
\usepackage{listings}
\makeatletter
\renewbibmacro*{textcite}{%
  \iffieldequals{namehash}{\cbx@lasthash}
    {\usebibmacro{cite:comp}}
    {\usebibmacro{cite:dump}%
     \ifbool{cbx:parens}
       {\bibclosebracket\global\boolfalse{cbx:parens}}
       {}%
     \iffirstcitekey
       {}
       {\textcitedelim}%
     \usebibmacro{cite:init}%
     \ifnameundef{labelname}
       {\printfield[citetitle]{labeltitle}}
       {\printnames{labelname}}%
     \addspace
     \ifnumequal{\value{citecount}}{1}
       {\usebibmacro{prenote}}
       {}%
     \mkbibsuperscript{\usebibmacro{cite:comp}}%
     \stepcounter{textcitecount}%
     \savefield{namehash}{\cbx@lasthash}}}
\makeatother

\usepackage{csquotes}
\usepackage{cleveref}
\usepackage{siunitx}
\usepackage{geometry}
\usepackage{multirow}
\usepackage[dvipsnames,table,xcdraw]{xcolor}
\definecolor{tumblue}{rgb}{0.5450980392,0.4901960784,0.6862745098}

\usepackage{xspace}
\usepackage{caption}
\usepackage{subcaption}
\usepackage{authblk}

\usepackage{marvosym}

\usepackage{sectsty} 

\sectionfont{\color{tumblue}\selectfont\sffamily}
\subsectionfont{\color{tumblue}\selectfont\sffamily}
\subsubsectionfont{\color{tumblue}\selectfont\sffamily}
\paragraphfont{\selectfont\sffamily}
\subparagraphfont{\color{gray}\selectfont\sffamily}

\definecolor{halfgray}{gray}{0.55}

\IfFileExists{./fonts/Heliotrope-Bold.otf}{
    \usepackage{fontspec}
    \setmainfont{Heliotrope}[Extension = .otf, UprightFont = *-Regular, ItalicFont = *-Italic, BoldFont=*-Bold, Path = ./fonts/]
}{
    \usepackage{biolinum}
    
}

\usepackage[labelfont=bf]{caption}
\usepackage[nonumberlist,acronym]{glossaries}
\input{acronyms}
\makenoidxglossaries 
\setacronymstyle{long-short}

 \usepackage{microtype}
 \usepackage{fontawesome}

\definecolor{linkcolor}{RGB}{0, 102, 204}

\newcommand{\gptfouro}{GPT-4o\xspace}

\newcommand{\claudethreefivesonnet}{Claude 3.5 Sonnet\xspace}

\newcommand{\llamathreetwoninety}{Llama 3.2-90B\xspace}

\newcommand{\geminipro}{Gemini Pro\xspace}
\newcommand{\llamavision}{Llama 3.2 90B Vision\xspace}

\newcommand{\macbench}{MaCBench\xspace}
\newcommand{\uspplot}{US Patent plots\xspace}
\newcommand{\chembench}{ChemBench\xspace}

%% file: acronyms.tex
\newacronym{llm}{LLM}{large language model}
\newacronym{ml}{ML}{machine learning}
\newacronym{agi}{AGI}{artificial general intelligence}
\newacronym{api}{API}{application programming interface}
\newacronym{mcq}{MCQ}{multiple-choice question}
\newacronym{rest}{REST}{representational state transfer}
\newacronym{orm}{ORM}{object relational mapping}
\newacronym{dai}{DAI}{daily allowed intake}
\newacronym{ghs}{GHS}{globally harmonized system of classification and labelling of chemicals}
\newacronym{who}{WHO}{World Health Organization}
\newacronym{nmr}{NMR}{nuclear magnetic resonance}
\newacronym{helm}{HELM}{Holistic Evaluation of Language Models}
\newacronym{smiles}{SMILES}{simplified molecular input line-entry system}
\newacronym{pca}{PCA}{principal component analysis}
\newacronym{iupac}{IUPAC}{International Union of Pure and Applied Chemistry}
\newacronym{json}{JSON}{JavaScript object notation}
\newacronym{rag}{RAG}{retrieval augmented generation}
\newacronym{ece}{ECE}{expected calibration error}
\newacronym{cot}{CoT}{chain of thought}
\newacronym{rlhf}{RLHF}{reinforcement learning with human feedback}
\newacronym{vllm}{VLLM}{Vision Large Language Model}
\newacronym{mof}{MOF}{metal-organic framework}
\newacronym{xrd}{XRD}{X-ray diffraction}
\newacronym{afm}{AFM}{atomic force microscopy}
\newacronym{ms}{MS}{mass spectrometry}
\newacronym{cif}{CIF}{Crytallographic Information Files}
\newacronym{qa}{QA}{Question-Answer Pairs}
\newacronym{mae}{MAE}{mean absolute error}

%% file: authors.tex
\credit{Nawaf Alampara}{1,1,1,0,1,1,0,0,1,0,1,1,0,1}
\credit{Mara~Schilling-Wilhelmi}{1,1,1,0,1,1,0,0,0,0,1,1,0,1}
\credit{Martiño~Ríos-García}{1,1,1,0,1,1,0,0,1,0,1,1,0,1}
\credit{Indrajeet~Mandal}{1,1,0,0,1,1,0,0,0,0,1,0,0,1}
\credit{Pranav~Khetarpal}{1,1,0,0,1,1,0,0,0,0,1,0,0,1}
\credit{Hargun~Singh~Grover}{1,1,0,0,1,1,0,0,0,0,1,0,0,1}
\credit{N.~M.~Anoop~Krishnan}{1,1,1,1,1,1,1,1,0,1,1,0,0,1}
\credit{Kevin Maik Jablonka}{1,1,1,1,1,1,1,1,0,1,1,0,1,1}

\author[1]{Nawaf~Alampara~\orcidlink{0009-0001-7697-7315}}
\author[1]{Mara~Schilling-Wilhelmi~\orcidlink{0009-0007-4392-5918}}
\author[1]{Martiño~Ríos-García~\orcidlink{0000-0003-1507-4048}}
\author[2]{Indrajeet~Mandal~\orcidlink{0000-0002-8808-4602}}
\author[3]{Pranav~Khetarpal~\orcidlink{0009-0001-8817-9134}}
\author[3]{Hargun~Singh~Grover}
\author[3,4, \Letter]{N.~M.~Anoop~Krishnan~\orcidlink{0000-0003-1500-4947}}
\author[1,5,6,7, \Letter]{Kevin~Maik~Jablonka~\orcidlink{0000-0003-4894-4660}}

\affil[1]{Laboratory of Organic and Macromolecular Chemistry (IOMC), Friedrich Schiller University Jena, Humboldtstrasse 10, 07743 Jena, Germany}
\affil[2]{School of Interdisciplinary Research, Indian Institute of Technology Delhi, Hauz Khas, New Delhi 110016, India
}
\affil[3]{Department of Civil Engineering, Indian Institute of Technology Delhi, Hauz Khas, New Delhi 110016, India
}
\affil[4]{Yardi School of Artificial Intelligence, Indian Institute of Technology Delhi, Hauz Khas, New Delhi 110016, India
}

\affil[5]{Center for Energy and Environmental Chemistry Jena (CEEC Jena), Friedrich Schiller University Jena, Philosophenweg 7a, 07743 Jena, Germany}
\affil[6]{Helmholtz Institute for Polymers in Energy Applications Jena (HIPOLE Jena), Lessingstrasse 12-14, 07743 Jena, Germany}

\affil[7]{Jena Center for Soft Matter (JCSM), Friedrich Schiller University Jena, Philosophenweg 7, 07743 Jena, Germany}

\affil[\Letter]{\texttt{krishnan@iitd.ac.in} and 
\texttt{mail@kjablonka.com}}

%% file: sections/introduction.tex
\section{Introduction}

The practice of science has always required assimilating and integrating diverse forms of information, from visual observations in the laboratory and measurements to theoretical frameworks and prior literature. While automation has traditionally excelled at repetitive tasks such as high-throughput experimentation,\autocite{Mahjour2023, Lu2023, Gesmundo2023, Wagen2022} capturing the fundamental characteristic of scientific work --- the ability to interpret and connect multiple modes of information flexibly --- has remained a central challenge for scientific discovery.

Recent advances in artificial intelligence, particularly in \glspl{llm}, have sparked renewed interest in developing more flexible computational systems for scientific workflows. These models can orchestrate specialized tools and combine general reasoning capabilities with domain-specific functions, suggesting a path toward more adaptable scientific automation.\autocite{ai4science2023impact, jimenez2024swebenchlanguagemodelsresolve, laurent2024labbenchmeasuringcapabilitieslanguage, miret2024llms, White2023, jablonka202314, ramos2024review} 
However, a fundamental challenge persists: bridging the gap between human scientists' natural ability to seamlessly integrate visual, numerical, and textual information and the current limitations of computational systems in processing these different data types. 
This gap becomes particularly apparent in tasks that require combining visual interpretation with scientific reasoning, such as analyzing spectroscopic data,\autocite{bushuiev2024massspecgymbenchmarkdiscoveryidentification} interpreting experimental setups,\autocite{Intelligent.com_2023} or evaluating safety conditions in laboratories.\autocite{Urbina_2022, campbell2023censoring}

Recent work has shown promising capabilities of \glspl{llm} in scientific tasks, from literature mining\autocite{schilling2024text, Polak2024, schillingwilhelmi2024using, D4DD00091A, dagdelen2024structured, caufield2023structured, skarlinski2024language, gupta2022discomat} and property prediction\autocite{Jablonka_2024,ramos2023bayesian, zhong2024benchmarking, xie2024fine, jablonka202314, kristiadi2024sober, gruver2024fine, alampara2024mattext} to experiment planning\autocite{boiko2023autonomous,darvish2024organa, bran2023chemcrow, swanson2024virtual}. Similarly, \glspl{vllm} have demonstrated increasing capabilities in general visual reasoning tasks\autocite{lu2022learnexplainmultimodalreasoning, gupta2024polymathchallengingmultimodalmathematical, cheng2024visionlanguagemodelsselfimprovereasoning, zou2024dynamathdynamicvisualbenchmark, shao2024visualcotadvancingmultimodal}. While recent benchmarks have evaluated either the scientific reasoning capabilities of language models\autocite{mirza2024large, Zaki_2024} or general multimodal abilities\autocite{lu2022learnexplainmultimodalreasoning, wang2024scibenchevaluatingcollegelevelscientific, gupta2024polymathchallengingmultimodalmathematical, zhang2024mathversedoesmultimodalllm}, a systematic evaluation of how these models handle the interplay of different modalities across the entire scientific process has been missing.  
This raises a crucial question: What are the limits of these models as copilots accelerating materials and chemistry research involving multimodal information extraction, simulations or experiments, and data analysis? While we have some understanding for text-only \glspl{llm}, we still have no understanding for \glspl{vllm} that can process images alongside text.

To address this gap, we present \macbench (materials and chemistry benchmark), a comprehensive benchmark that evaluates multimodal capabilities across three fundamental pillars of the scientific process: information extraction from the literature, experiment execution, and data interpretation. By focusing on these pillars, we can assess models' abilities across the full spectrum of scientific tasks, from understanding published results to executing and interpreting new experiments.
Our benchmark is distinctively designed to not only measure performance but also to uncover the underlying failure modes of current models systematically. Through carefully constructed ablation studies, we investigate how performance varies across different modalities, levels of domain expertise required, reasoning complexity, and the distance to the training data corpus. This systematic approach allows us to test the hypothesis that current models might rely on superficial pattern matching rather than deeper scientific understanding.
Our results reveal that while models can handle certain modalities individually, they often fail when tasks require flexible integration of information types---a core capability required for scientific work. For instance, models might correctly perceive information but struggle to connect these observations in scientifically meaningful ways. 

These insights have important implications for developing AI-powered scientific assistants and self-driving laboratories. Our results highlight the specific capabilities needing improvement for these systems to become reliable partners in scientific discovery. They also suggest that fundamental advances in multimodal integration and scientific reasoning may be needed before these systems can truly assist in the creative aspects of scientific work.

%% file: sections/results.tex
\section{Results}

\subsection{The \macbench framework}
\begin{figure}
    \centering
    \includegraphics[width=\textwidth]{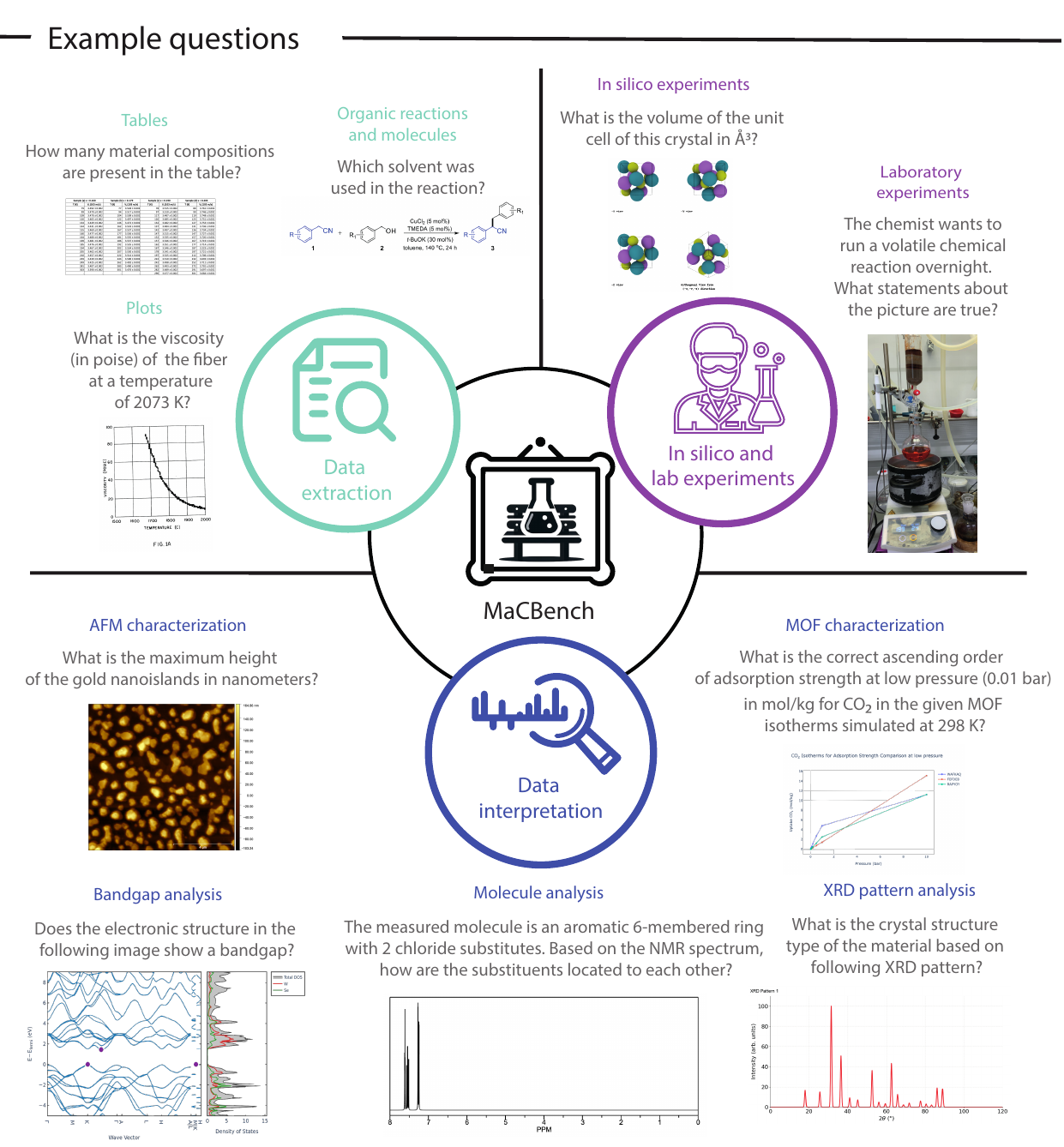}
    \caption{\textbf{Overview of the \macbench framework, covering the multimodal chemistry and materials science research life cycle.} The framework evaluates \gls{vllm} performance across three key domains: data extraction (teal), in silico and laboratory experiments (purple), and data interpretation (pink). 
    The benchmark includes diverse tasks spanning tables, plots, organic chemistry diagrams, crystal structures, \gls{afm} imaging, spectroscopy, and materials characterization. Each task requires domain-specific visual understanding and scientific reasoning, from extracting numerical values to analyzing complex experimental setups and interpreting spectroscopic data. We use icons created by Rainy Ting (on \url{svgrepo.com}).}
    \label{fig:overview}
\end{figure}

Our benchmark design is guided by the observation that scientific work requires not only access to multiple modalities of information but also the ability to flexibly integrate them. 
To probe these capabilities of \glspl{vllm} meaningfully --- rather than creating artificial question-answer-based challenges --- we focus on tasks that mirror real scientific workflows, from interpreting scientific literature to evaluating laboratory conditions and analyzing experimental data (see \Cref{fig:overview}). 
This approach allows us to evaluate the models' ability to process different types of information and their capacity to use this information to support scientific discovery.
To assess performance in a broad range of settings, we rely on both images we mined from patents but also some we generated from scratch. 

\begin{figure}

    \centering
    \includegraphics[width=1.0\textwidth]{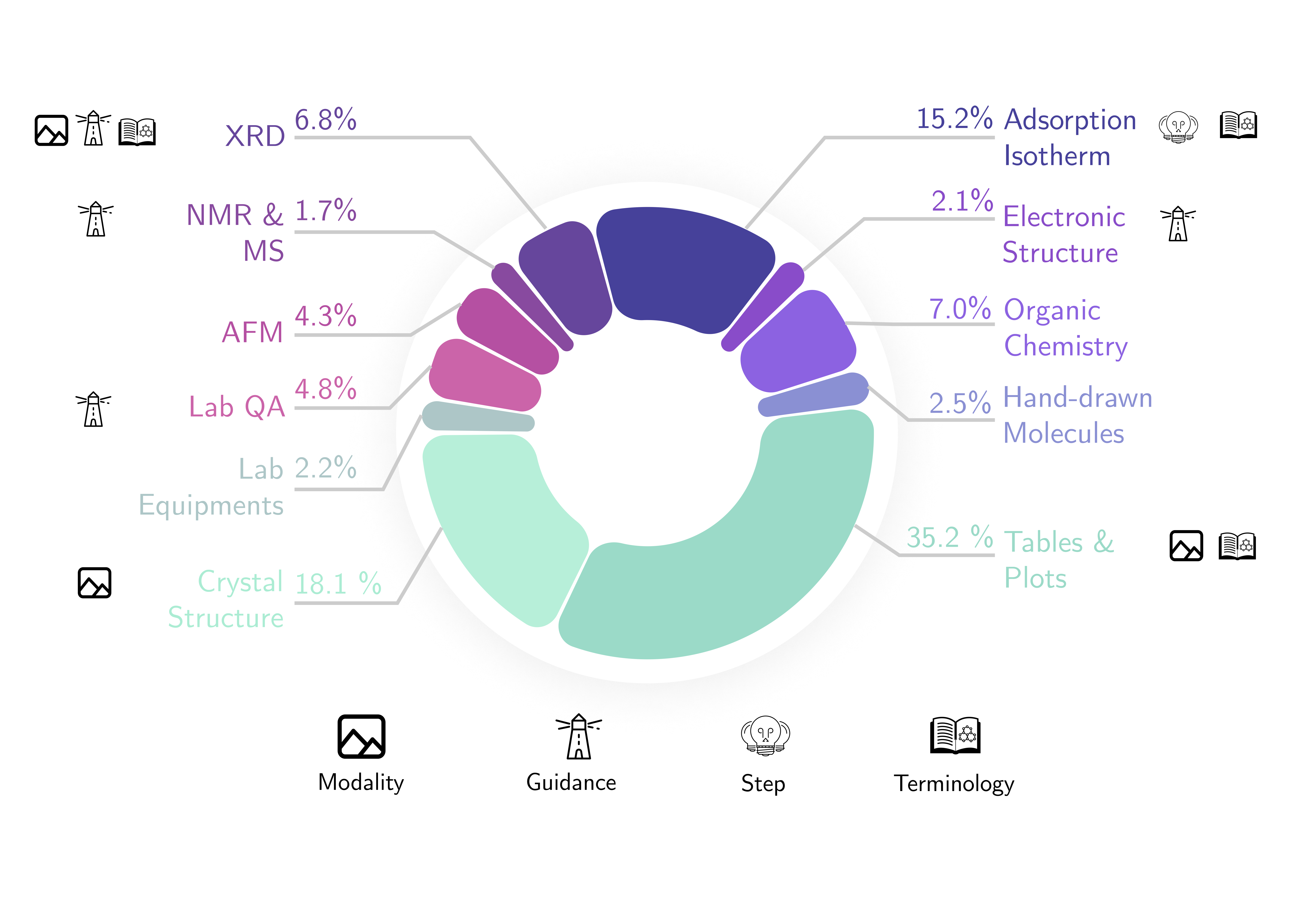}
    \caption{\textbf{Distribution of tasks in the \macbench dataset.} \macbench comprises nine distinct task categories with their respective proportions, ranging from Tables \& Plots (\SI{35.2}{\percent}) to \gls{ms} \& \gls{nmr} analysis (\SI{1.7}{\percent}).
    Each segment is annotated with relevant icons indicating the ablations we conducted on those tasks: modality understanding (image icon), guidance requirements (lighthouse icon), reasoning steps (lightbulb icon), and terminology complexity (book icon). The chart illustrates the benchmark's comprehensive coverage of chemistry and materials tasks.}
    \label{fig:macbench_compositions}
\end{figure}

The benchmark is structured around three key aspects that form the basis of many scientific workflows: information extraction, in silico or laboratory experiments, and data interpretation. Within each pillar, we include tasks spanning various scientific activities (see \Cref{fig:macbench_compositions}). The information extraction pillar analyzes the performance in parsing scientific literature, including extracting data from tables and plots and interpreting chemical structures. The experiment execution pillar evaluates the models' ability to understand laboratory safety, identify equipment, assess safety conditions, and understand crystal structures (as potential simulation artifacts). 
The data interpretation pillar tests models' capability to analyze various types of scientific data, from spectral analysis to electronic structure interpretation.

\subsection{Performance landscape}
\begin{figure}[ht]
    \centering
    \includegraphics[width=\textwidth]{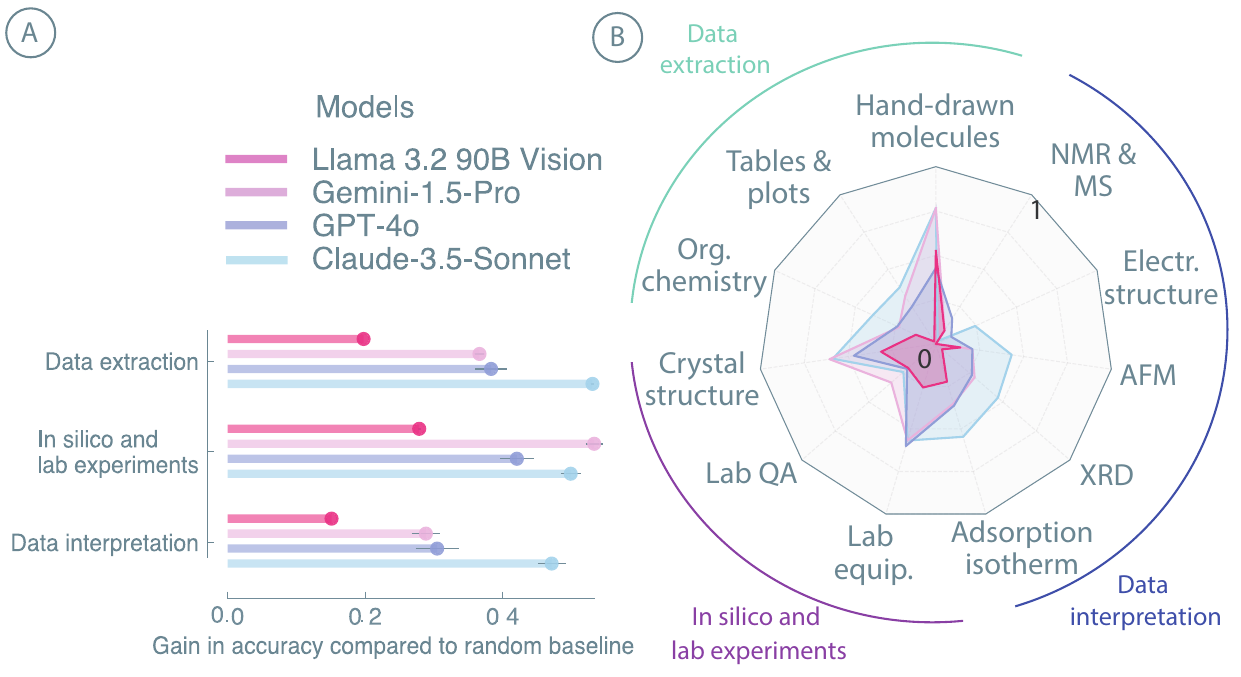}
    \caption{\textbf{Performance of frontier \glspl{vllm}.} \textbf{a.} Accuracy gains compared to random baseline across three core scientific tasks, showing varied performance of \claudethreefivesonnet, \gptfouro, \geminipro, and \llamavision in averaged across all task in the three focus areas of \macbench: data extraction, experimental understanding, and interpretation tasks. We show the performance as the fraction of correctly answered questions relative to a random baseline. A performance of 0 means that the model is indistinguishable from random guessing. The error bars indicate the standard deviation of the fraction of correctly answered questions over five different runs. 
    \textbf{b.} Radar plot demonstrating the relative model performance across ten specialized scientific domains. Again, we show the fraction of correctly answered questions relative to a random baseline (the plots without the normalization are shown in \Cref{fig:overall-performance_unnormalized}). We can observe substantial differences in performance across topics.}
    \label{fig:overall-performance}
\end{figure}

There is significant variation in model performance across different task types and modalities (\Cref{fig:overall-performance}, see \Cref{tab:domain_topics} for detailed descriptions of all tasks). However, when averaged over different tasks, \claudethreefivesonnet is the leading model on all three task families.
In addition, it is interesting to note that the models do not fail at one specific part of the scientific process but struggle in all of them, suggesting that broader automation is not hindered by one bottleneck but requires advances on multiple fronts. 
Interestingly, even for the first step of the scientific process --- data extraction --- some models do not perform much better than random guessing (e.g., \llamavision in \Cref{fig:overall-performance}).
Current systems tend to perform best on multiple-choice-based perception tasks (e.g., lab equipment and hand-drawn molecules in \Cref{fig:overall-performance}).

\paragraph{Data extraction}

Interestingly, our analysis shows that the first step of the scientific workflow, data extraction, already poses considerable challenges for the models we tested. This is particularly the case for extracting science-specific data, for instance, about organic reactions and molecules. While the best models perform well at extracting information about reaction diagrams, they fail to correctly describe the relationship between isomers (see \Cref{fig:organic_performance}). As discussed below, this is likely caused by models struggling with spatial reasoning.
In addition, even the extraction of compositions from tables still shows room for improvement for the \glspl{vllm} we tested (average accuracy of 0.53), performing not distinguishable from random guessing for \llamavision.

\paragraph{In silico and lab experiments}
A similar variance in performance is observed for tasks related to the execution of laboratory or in silico experiments. 
While models show good performance in recognizing laboratory equipment (average accuracy of 0.77), reasoning about lab scenarios, for example, comparing the safety hazards of two similar lab setups, shows low performance (average accuracy of 0.46). 

The disparity between equipment identification and safety assessment performance suggests that while models can learn to recognize standard laboratory equipment, they still struggle with the more complex reasoning required for safe laboratory operations, questioning their ability to assist in real-world experiment planning and execution.
This finding also implicates that current models cannot bridge gaps in tacit knowledge frequently discussed in biosafety scenarios.\autocite{barrett2024benchmarkearlyredteam, sandbrink2023artificialintelligencebiologicalmisuse} 

Also, the interpretation of crystal structure renderings, a crucial step for in silico experiments, shows performance that is indistinguishable from random guessing in some cases, such as the assignment of space groups (see \Cref{fig:cif_performance}). 

\paragraph{Data interpretation}
Interpreting experimental results often proves challenging to all models, including \claudethreefivesonnet. While most models can interpret capacity values (average accuracy of 0.59), compare Henry constants (average accuracy of 0.83) from MOF isotherms, or interpret amorphous or crystalline systems from XRD with acceptable performance (average accuracy of 0.69), they struggle to interpret \acrfull{afm} images (average accuracy of 0.24) and often fail with tasks involving measurements like width and length (despite the presence of clear legends). 
They also fail to reliably interpret \gls{ms} and \gls{nmr} spectra (average accuracy of 0.35) or to make inferences on \gls{xrd} pattern. 
In the latter case, it is particularly striking that while some models perform very well in identifying the positions of the most intense reflections, they perform poorly in determining relative orderings, crucial for interpreting \gls{xrd} patterns.

\subsection{Understanding model limitations}

\begin{figure}
    \centering
    \includegraphics[width=1\linewidth]{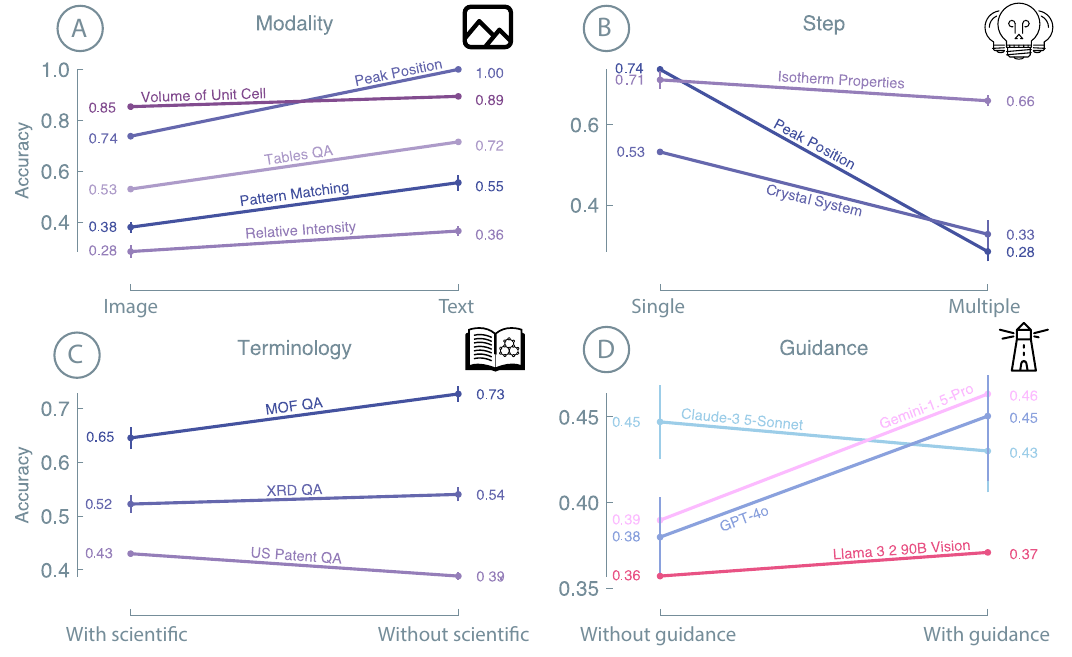}
    \caption{\textbf{Ablation study results across four key dimensions of \glspl{vllm} performance in chemistry and materials science tasks.} \textbf{a.} Modality analysis compares performance between image-only and text-only inputs across different task types, with typically higher performance when the same information is shown in text form. \textbf{b.} Step complexity analysis demonstrates performance degradation as tasks require multiple reasoning steps. \textbf{c.} Terminology impact shows how scientific language specificity affects model accuracy, comparing performance with and without domain-specific terminology. We found the behavior on US Patent QA to be mostly due to the sensitivity of \geminipro to the prompt template (see \Cref{sec:prompt_fragility}) \textbf{d.} The guidance study compares performance across different \glspl{vllm} with and without additional task guidance, revealing model-specific sensitivity to prompting strategies. For each task, we calculated the mean score and standard deviation across five independent runs. To summarize performance across models, we averaged the mean scores and standard deviations for each task. For combined tasks (e.g., \enquote{XRD QA}, \enquote{Isotherm QA}, \enquote{Tables QA}), we employed a two-step averaging process.
     For each model, we averaged the scores and standard deviations across the sub-tasks. We then averaged these model-specific averages across all models to obtain the final mean score and standard deviation for the combined task.
    For guidance analysis, performance was measured as the mean score across five independent runs, and the variability was quantified using the standard deviation of those runs. To obtain an overall measure of performance and variability for each side (with and without guidance), we calculated the mean score and the mean standard deviation across all tasks within each side.}
    \label{fig:ablation_lines}
\end{figure}

To further understand the failure modes of \glspl{vllm}, we designed a comprehensive suite of ablation studies.
Our approach isolates specific aspects of scientific tasks, from the complexity of reasoning required to how information is presented. 
We probe two distinct categories of limitations (\Cref{fig:ablation_lines}): first, core reasoning limitations that appear fundamental to current model architectures or training approaches or datasets, and second, sensitivities to inference choices.

\paragraph{Core Reasoning Limitations}
Some limitations appear intrinsic to current model architectures and are unlikely to be overcome regardless of how tasks are presented or prompted.  These fundamental constraints manifest in three key areas.

\subparagraph{Spatial reasoning}
While one might expect \glspl{vllm} to excel at processing spatial information, our results reveal significant limitations in this capability.
For example, while models achieve high performance in matching hand-drawn molecules to \gls{smiles} strings (average accuracy of 0.80, four times better than baseline), they perform almost indistinguishably from random guessing for naming the isomeric relationship between two compounds (e.g., enantiomer, regioisomer, average accuracy 0.24 only 0.1 higher than the baseline accuracy ) and when assigning stereochemistry (average accuracy of 0.24, baseline is 0.22).
Similarly, models perform well in simple perception tasks on crystal structures (e.g., counting the number of different species, average accuracy of 0.85) but struggle at assigning the crystal system (average accuracy of 0.55) or space groups (average accuracy of 0.45).

These striking performance drops for tasks requiring spatial reasoning suggest that current \glspl{vllm} cannot reliably be used for any tasks requiring this capability --- even though this might be one of the most intuitive use cases of these models.

\subparagraph{Synthesis across modalities}
Given that models consume visual and textual input in seemingly similar ways, one might expect that the same information is processed in the same way regardless of how it is presented to the model.

To probe the ability of models to integrate information across modalities, we presented identical information in both text and image. 
In \Cref{fig:ablation_lines}, we find that for all tasks where we show the same information as images and text, the performance in the text modality is better than when the information is provided as an image.
A striking example emerges when identifying the peak position in XRD. Models shows nearly a 35\% increase in the performance when presented with the same peak positions in text against visually showing the peaks.
Even when calculating the volume of crystal structures, models show four percentage point difference in performance when presented with the structural information in visual form (unit cell parameters shown in the image) and textual form (unit cell parameters shown in text). 
These results suggest that current models have not yet developed robust strategies for cross-modal information synthesis.

\subparagraph{Multi-step reasoning}
Motivated by the fact that the overall performance analysis indicated that perception tasks tended to be best, we designed experiments in which we probe, with the same inputs\autocite{mccoy2023embersautoregressionunderstandinglarge}, the performance on very similar tasks but requiring different numbers of reasoning steps (or different numbers of tool calls when implemented in an agentic framework).

Our analysis reveals consistent degradation in performance as tasks require more reasoning steps. \Cref{fig:ablation_lines} shows that in all our experiments, the tasks requiring multiple steps perform significantly worse than those requiring only one step.
For instance, in \gls{xrd} pattern analysis, models perform significantly better at identifying the highest peak than at ranking relative peak intensities (average accuracy of 0.74 for identification of the highest peak against 0.28 for ranking). Similarly, for the interpretation of adsorption isotherms, accuracy in finding the highest value notably exceeds performance in ordering multiple values. 
This pattern suggests fundamental limitations in chaining logical steps, a crucial capability for scientific reasoning.

\paragraph{Sensitivity to inference choices}
While addressing these core limitations will require novel training approaches, we also identified several factors that significantly influence model performance through inference choices rather than fundamental capabilities.
Those factors present an actionable way to improve the performance of current systems directly without retraining them.

\subparagraph{Scientific terminology}
One might hypothesize that models struggle with some tasks because they are unfamiliar with the scientific terminology used in the questions. \Cref{fig:ablation_lines} shows that removing scientific terminology improves performance across some tasks, including the analysis of adsorption isotherms of \gls{mof}, \gls{xrd} pattern interpretation. Similarly, using \gls{iupac} names instead of \gls{smiles} notation for chemical compound identification leads to better results. 
This suggests models might be overly sensitive to specific technical vocabularies rather than understanding underlying concepts.
In fact, some models like \geminipro (and the surrounding refusal mechanisms) are very sensitive to the exact wording of the prompt. In \Cref{sec:prompt_fragility}, we show that for some questions, large variations in performance can be due to apparently minor changes in prompt wording, such as replacing the word \enquote{image} with \enquote{diagram,} \enquote{plot,} \enquote{figure,} \enquote{photograph,} or even omitting it entirely. 
 
\subparagraph{Guidance following} 
Given that chemists receive instructions on interpreting various experimental characterizations, we hypothesized that similar guidance might also help the models perform better on such tasks. 
Interestingly, adding step-by-step instructions improves performance for most models in spectral analysis, electronic structure interpretation, and \gls{xrd} pattern matching—with the notable exception of \claudethreefivesonnet, whose performance does not improve when provided with guidance. 
This variation in response to instruction suggests different underlying approaches to problem-solving across models.

\subsection{Performance as a function of frequency on the internet}
\begin{figure}
    \centering
    \includegraphics[width=\textwidth]{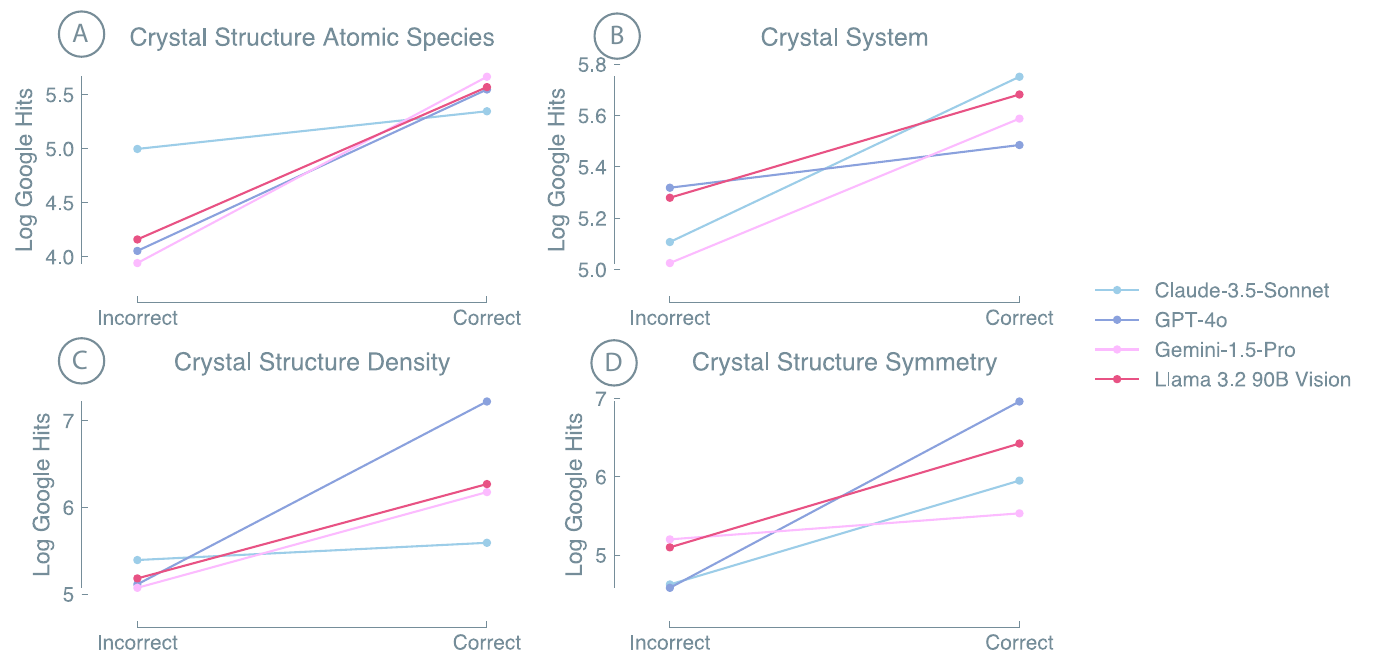}
    \caption{\textbf{\gls{vllm} performance as a function of number of search hits.} The plots compare four leading \glspl{vllm} across different crystallographic tasks: \textbf{a.} atomic species identification, \textbf{b.} crystal system classification, \textbf{c.} density calculation, and \textbf{d.} crystal symmetry determination. For each property, the log-scale Google hit counts are plotted against the correctness of model responses, revealing correlations between answer accuracy and the prevalence of information in online sources. Higher hit counts for correct answers suggest models may not solely rely on reasoning in their responses to crystal structure analysis tasks.}
    \label{fig:search_hit_correlation}
\end{figure}

The varying impact of guidance across models led us to investigate whether models truly engage in scientific reasoning or primarily match patterns from their training data.\autocite{mccoy2023embersautoregressionunderstandinglarge} 
To probe this question, we measured the number of Google search results for various crystal structures as a proxy for the frequency of those structures in the training corpus (\Cref{fig:search_hit_correlation}).

Our analysis reveals a striking correlation between the prominence of crystal structures on the Internet and task performance. \Cref{fig:search_hit_correlation} shows that for all cases in our benchmark, the structures for which the models solve the tasks are more prominent on the Internet.
This suggests that models might rely more on pattern matching than genuine scientific reasoning. Interestingly, we observe this effect even for tasks that depend solely on perception, such as counting the number of distinct atomic species.

\subsection{Toward robust multimodal assistants}

Our analysis reveals the promise and limitations of state-of-the-art \glspl{vllm} in scientific tasks. Compared to text-only benchmarks such as the one of \textcite{mirza2024large}, we observe substantially higher performance variability across tasks, suggesting that multimodal systems are more fragile than \glspl{llm}. 
This fragility manifests in several ways: the striking performance gap between visual and textual representations of identical information indicates incomplete integration of modalities, while the strong correlation between model performance and the Internet presence of specific crystal structures raises questions about true reasoning capabilities versus pattern matching. 
The sensitivity to prompting choices (see \Cref{sec:prompt_fragility}) and the counterintuitive finding that guidance can degrade performance for top models further underscore reliability concerns. 
However, our findings also point to actionable paths forward. 
Many observed limitations, particularly in spatial reasoning, could potentially be addressed through synthetic training data generation. 
When pursuing such approaches, we recommend incorporating generalization tests (e.g., evaluating spatial reasoning on larger compounds than those in training\autocite{anil2022exploringlengthgeneralizationlarge}) to ensure robust capability development. 
Furthermore, the significant performance differences between modalities suggest opportunities for improved training strategies, such as incorporating modality transformation tasks (e.g., automated conversion between spectral data representations). 
These targeted interventions could help bridge the gap between current capabilities and the needs of scientific workflows.

%% file: sections/conclusions.tex
\section{Conclusions}
Scientific reasoning is fundamentally a multimodal process. Current vision-language models show promising capabilities in simple cases, such as identifying laboratory equipment or extracting explicit numerical values from plots. For standardized representations like \gls{smiles} notations or simple spectra, models can even achieve high accuracy in information extraction. 
However, model performance becomes unreliable when tasks require the integration of visual and conceptual understanding---as in complex laboratory safety assessments or crystal structure analysis.

Through careful ablation studies, we found that despite their impressive scale and training, current \glspl{vllm} require significant improvement in their vision modality as they seem to perform drastically better when the same information is shown in text instead of as an image. Moreover, the models seem to rely on pattern matching rather than developing robust scientific understanding. This becomes particularly evident in the observation that model performance correlates strongly with online prominence.

Yet, our benchmark also demonstrates the remarkable progress in AI systems' ability to process scientific information, with (almost) perfect performance achieved in several tasks. 
The observation that performance can be improved through careful terminology choice and task guidance (though with model-specific variations) suggests practical paths forward. More broadly, our findings indicate that advancing AI in science requires not just model improvements but also better ways of representing scientific knowledge---particularly in addressing the observed gaps in spatial reasoning and cross-modal integration capabilities. 

While current \glspl{vllm} cannot yet serve as autonomous scientific reasoners, they show promise as assistive tools when their limitations are well understood and their deployment is carefully structured around their demonstrated strengths. 
As we continue to develop these systems, our work suggests that advancing from pattern matching---demonstrated by the strong correlation between model performance and internet presence of crystal structures—--to true scientific reasoning may require fundamental advances in both training data curation and model architectures that can better handle spatial relationships and cross-modal information synthesis.

%% file: sections/methods.tex
\section{Methods}
\begin{figure}
    \centering
    \includegraphics[width=.7\linewidth]{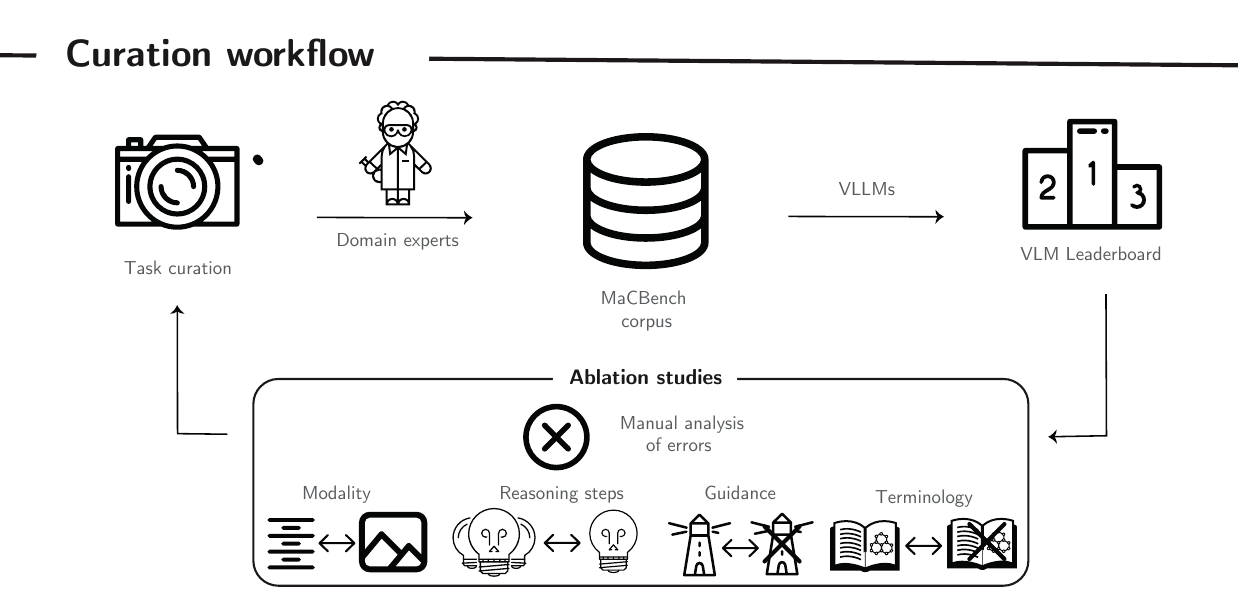}
    \caption{\textbf{The \macbench curation workflow.} Tasks are initially collected and curated through manual selection, followed by validation by domain experts in chemistry and materials science. The validated tasks form the \macbench corpus, which is used to evaluate various \glspl{vllm}, resulting in a performance leaderboard. Ablation studies are conducted through manual error analysis focusing on four key aspects: modality understanding, reasoning steps, guidance requirements, and terminology usage. Results from these analyses feed back into the task curation process, enabling continuous benchmark refinement.}
    \label{fig:curation_workflow}
\end{figure}

Our question curation and model evaluation methodology leverages the \chembench framework.\autocite{mirza2024large}
For curation, we manually sourced questions and then created ablations based on error analyses to systematically understand failure modes (\Cref{fig:curation_workflow}). For most tasks, we created new images, e.g., by building and photographing lab setups or by plotting experimental data.
Similar to \textcite{mirza2024large}, all questions have been reviewed by multiple scientists before being entered into the corpus.
In the curation process, we also recorded tolerances for each question. That is, for each numerical answer, we recorded windows within which an answer would still be deemed correct to account for natural uncertainties and noise.

\paragraph{Dataset} Our questions in the dataset are stored in an extended BigBench format \autocite{srivastava2023beyond}. Each question, along with its corresponding base64-encoded image, is stored in separate JSON files. To prevent potential data leakage during future model training, the BigBench canary string is included in each file. Our pipeline employs a robust templating system, allowing for the dynamic insertion of multiple images and other text template elements into questions using placeholders. This enables our pipeline to interleave images directly into question prompts in designated locations.

All questions in our benchmark contain pairs of images and text-based questions. Only some ablation experiments (that are specifically highlighted) contain only text information. 

\paragraph{Evaluation} We employ \chembench's prompt templates for instruction-tuned models, which also impose specific response formats on the models. The parsing workflow, also based on \chembench, utilizes regex-based functions to extract answers from various scientific notations, handling both multiple-choice responses and numerical values. The regex-based parsing is backed up with an \gls{llm} extractor (e.g., \claudethreefivesonnet) for cases where standard parsing fails. We included the encoded images in the prompt. We used the default quality setting for each provider. That is, for \geminipro images will be automatically scaled up or down to fit into the allowed range ( $768\times768$ - $3072\times3072$), while for \claudethreefivesonnet if the image’s long edge is more than 1568 pixels it is scaled down.
For \llamavision, an \gls{api} error will be raised if the images are bigger than allowed. For \gptfouro, the default configuration is set to \enquote{auto}, meaning that the quality of the images is automatically selected by the \gls{api}. For low-resolution images, they are set to $512\times512$ pixels. For the high-resolution mode, the model first sees the $512\times512$ image, then crops the image into $512\times512$ pixels tiles that are studied individually.

\paragraph{Refusal}
We implement a comprehensive framework combining regular expression-based detection from \gls{llm} Guard and a fine-tuned BERT model \autocite{distilroberta-base-rejection-v1} to identify potential \gls{llm} refusals. This detection pipeline was integrated into our evaluation pipeline, enabling pre-scoring refusal checks. To mitigate refusals, we implemented an interval-based retry mechanism, re-querying the LLM up to $n$ times until a non-refusal response was obtained. For our runs we retry for a maximum of five times. A count on the refusal by different models is shown in \Cref{tab:refusal_table}.

\paragraph{Relative performance} To account for the fact that for \glspl{mcq} a non-zero performance can be achieved that depends on the number of options, we report the metrics in the main text as performance gains over the performance this random baseline would achieve: 

\begin{equation}
    \mathrm{acc}_\mathrm{rel} = \mathrm{acc} - \mathrm{acc}_\text{baseline}
\end{equation}

\paragraph{Correlation of performance with the number of search results} For analyzing the correlation between the performance of the models and the prominence of the web, we used the total number of results for querying the common name of crystal structures returned by the Serp API.

%% file: sections/data_availability.tex
\section*{Data and code availability}

To facilitate the benchmarking and reproducibility of our work, we have provided the datasets used in this work on Hugging Face (\href{https://huggingface.co/datasets/jablonkagroup/MaCBench}{DOI: 10.57967/hf/4611} and \href{https://huggingface.co/datasets/jablonkagroup/MaCBench-Ablations}{DOI: 10.57967/hf/4612} ). \autocite{lab_of_kevin_jablonka_at_uni_jena_2025,lab_of_kevin_jablonka_at_uni_jena_2025_}

The code for running the benchmark is available at \url{https://github.com/lamalab-org/chembench/} and archived on Zenodo (\href{https://doi.org/10.5281/zenodo.14935487}{DOI: 10.5281/zenodo.14935487}). Instructions for running the benchmark can be found at \url{https://lamalab-org.github.io/chembench/getting_started/#how-to-benchmark-on-multi-modal-tasks}.

%% file: sections/acknowledgment.tex
\section*{Acknowledgments}
This work was supported by the Carl Zeiss Foundation, and a \enquote{Talent Fund} of the \enquote{Life} profile line of the Friedrich Schiller University Jena. A grant from OpenPhilanthropy additionally supported parts of the work. 

In addition, M.S-W.'s work was supported by Intel and Merck via the AWASES programme. 

K.M.J.\ is part of the NFDI consortium FAIRmat funded by the Deutsche Forschungsgemeinschaft (DFG, German Research Foundation) – project 460197019. 

N.M.A.K.\ acknowledges the Google Research Scholar Award, the Alexander von Humboldt Foundation for funding support, and the HPC IIT Delhi for computational and storage resources. 

We thank Bastian Rieck for developing the \LaTeX-credit package (\url{https://github.com/Pseudomanifold/latex-credits}). We also thank Kristin Schreyer for helping in collecting the pictures for the Lab QA task and Nitya Nand Gosvami for providing the \gls{afm} images.

%% file: sections/competing_interests.tex
\section*{Competing interests}

K.M.J.\ has been a paid contractor for OpenAI (as part of the red teaming network).

%% file: appendix.tex
\section{Appendix}

\subsection{Desired properties of a chemistry and materials based multimodal benchmark} \label{sec:desired-properties}

\begin{itemize}
    \item \emph{Evaluation of the cognitive abilities of \glspl{vllm}}. The main requirement of a benchmark is to evaluate the performance of the current leading models in a set of robust, extensive, and representative tasks.
    
    \item \emph{Generalization on all real-world problems}. For fields such as chemistry or materials science, \glspl{vllm} are intended to help the scientists in their daily tasks, going from lab safety assistant to assisting in the planning and interpretation of the experimental work.
    
    \item \emph{Help to identify the limitations of the models}. To make future \glspl{vllm} more useful to the scientists, the benchmarks must identify current limitations and light the path to more practical models.
    
    \item \emph{Highlight strengths of the models}. Many of the current capabilities of \glspl{vllm} are still undiscovered. Showing light on these capacities can increase the usefulness of the current models.
    
    \item \emph{Image-text integration}. A key indicator of the performance of \glspl{vllm} is how well they can join and understand image and text inputs to produce meaningful outputs.
    
    \item \emph{Evaluation of the performance in noisy images}. To test the models' performance in complex tasks, include out-of-distribution tasks that evaluate the models' robustness against noise and atypical data.
    
    \item \emph{Task versatility}. Include tasks that include the possible scientific scenarios; these can include visual reasoning, visual data extraction, or visual interpretation.
\end{itemize}

\subsection{Related work}

The rapid development of \glspl{vllm},\autocite{liu2023visualinstructiontuning, gpt_4o, geminiteam2024gemini15unlockingmultimodal, claude_35} has led to the publication of numerous benchmarks focused on some domains such as the medical,\autocite{jeong2024medicaladaptationlargelanguage, } math,\autocite{gupta2024polymathchallengingmultimodalmathematical, zhang2024mathversedoesmultimodalllm, zou2024dynamathdynamicvisualbenchmark} general science,\autocite{li2024mmscidatasetgraduatelevelmultidiscipline, liang_etal_2024_scemqa} or general knowledge benchmarks.\autocite{yue2024mmmumassivemultidisciplinemultimodal, chia2024puzzlevqadiagnosingmultimodalreasoning, shao2024visualcotadvancingmultimodal, roberts2024image2structbenchmarkingstructureextraction, zhang2024mmllmsrecentadvancesmultimodal, cheng2024visionlanguagemodelsselfimprovereasoning}
In addition, some interesting benchmarks have been published focusing on chemistry, materials science, and related fields. Therefore, \textcite{laurent2024labbench} created a benchmark to evaluate \gls{llm}-powered agents. In the benchmark, they defined some tasks as multimodal images and tables, which evaluate the agents' capabilities in biological settings. \textcite{li2024mmscidatasetgraduatelevelmultidiscipline} created a broad scientific benchmark by extracting figures from some open-source general science journals and prompting the models with questions about them. Thus, the authors designed different visual tasks to evaluate its ChemVLM model and enhance their textual benchmark.\autocite{zhang2024chemllmchemicallargelanguage} 
\textcite{roberts2024scifibenchbenchmarkinglargemultimodal} created a benchmark focused on evaluating the interpretation and understanding of different scientific figures. 
Similarly, \textcite{khalighinejad2024matvixmultimodalinformationextraction} built a benchmark that is specifically focused on evaluating the data extraction capabilities of \glspl{vllm} in extracting polymers data from full scientific articles.
While the tasks and areas studied by the previous benchmarks reveal important insights, we target the focus of \macbench on the uncovered areas and tasks, such as Lab Safety, to fill the gaps in our comprehension of the models' capabilities in chemistry and materials science.

\subsection{Scoring methodology}

For \glspl{mcq}, a task is considered correct if the Hamming loss is zero, meaning the predicted answer exactly matches the ground truth. For numeric questions, a response is deemed correct if the \gls{mae} falls within a specified tolerance. The default tolerance is 1\%, but for certain question types, such as CIF-Density, CIF-Volume, and some US Patent questions, the tolerance is up to 5\% (the tolerance is defined in the curation process). 
Each correct task receives a score of 1, while incorrect tasks receive a score of 0.
  
\paragraph{Overall \& Topic Scores}
The overall score is calculated as the total number of correct tasks divided by the total number of questions, excluding ablation tasks. Topic-wise scores are computed similarly, with the total number of correct answers in a topic divided by the total number of questions in that topic. For ablation tasks, the same topic-wise scoring method is applied. When tasks or models are combined, their respective scores and standard deviations are averaged. 
 
 \paragraph{Baseline}
The random baseline is established by randomly selecting one answer from the available options for \glspl{mcq}. 
For example, if there's a \glspl{mcq} chemistry question asking "Which element has the highest electronegativity?" with options A) Fluorine, B) Oxygen, C) Nitrogen, and D) Chlorine, the baseline would simply pick one letter randomly (e.g., "C"). 
For numeric questions, we use the mean of all target values within a topic as the prediction.
For instance, if there are multiple questions asking for the number of atoms in CIF with answers like 6, 12, and 15, the baseline would calculate the average of all these values (11) and use that as its prediction for every numeric question in that topic.
 
 The entire benchmark is run five times, and the standard deviation of the overall and topic-wise scores is used as the error bar to account for variability.

\subsection{Tasks in the \macbench corpus}

To unveil the proficiency of the models, we carefully designed a set of specific tasks that we consider essential parts of the scientific workflow in the chemical sciences. \Cref{tab:domain_topics} include the name, number of questions, and descriptions for all the main tasks in the \macbench corpus.

\begin{table}
    \caption{\textbf{Number of questions and description of all the tasks in the \macbench corpus.} We grouped tasks in themes corresponding to typical tasks in the scientific workflow in the chemical sciences. Those groups correspond to the ones shown in the radar plots. All tasks shown in this table consist of an image shown alongside a question in text form.}
    \resizebox{\textwidth}{!}{
    \input{tables/description_table.tex}
    }
    \label{tab:domain_topics}
\end{table}

\paragraph{Model performance on the main tasks}

As mentioned in the main text, we evaluated some leading \glspl{vllm}. 
\Cref{tab:performance_table} collects the overall performance of the models along the different tasks. In that table, we also include the random baseline results, which are used as the base for the overall performance figure of the main text (see \Cref{fig:overall-performance}). 

\begin{figure}[ht]
    \centering
    \includegraphics[width=\textwidth]{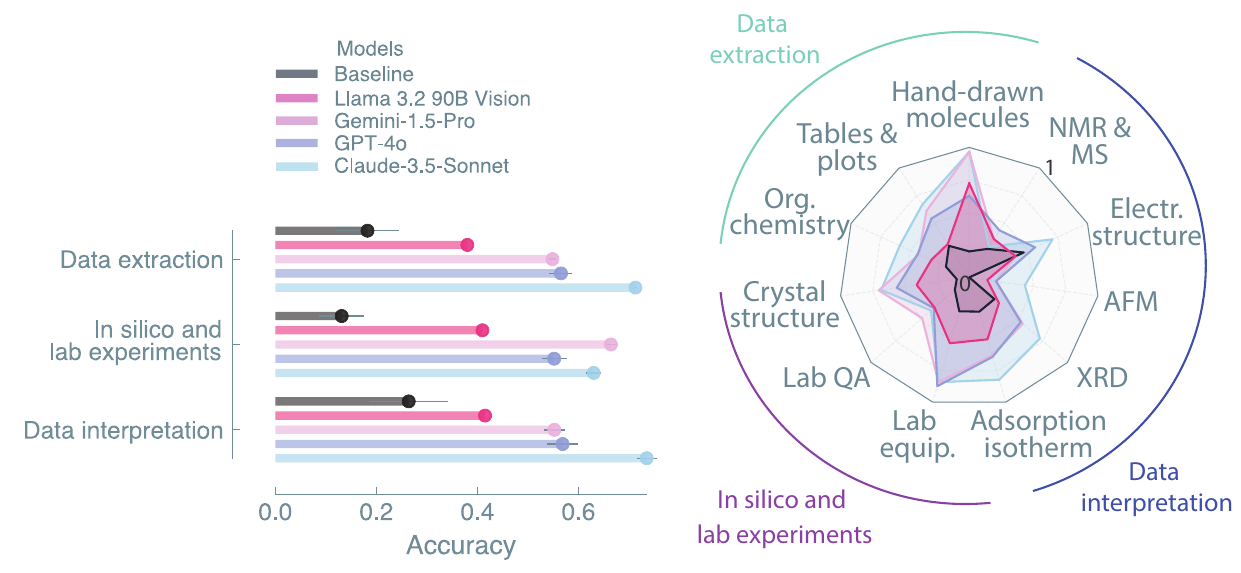}
    \caption{Performance of frontier vision-language models across scientific tasks, organized by the three pillars of the scientific process: information extraction, experiment execution, and data interpretation. While models show strong performance in certain basic tasks, their capabilities vary significantly when deeper scientific reasoning is required. The error bars in the bar plot indicate the standard deviation of a fraction of correctly answered questions over five different runs.}
    \label{fig:overall-performance_unnormalized}
\end{figure}

Similarly, to better illustrate the overall results, \Cref{fig:overall-performance_unnormalized} visually describes the performance of the models along all the \macbench tasks, including the random baseline as the fifth model.

\begin{table}
    \caption{\textbf{Absolute performance of the models on the \macbench corpus classified by the three pillars considered in \macbench.} Note that in this table, the random baseline is included as a model.}
    \resizebox{\textwidth}{!}{
    \input{tables/main_results.tex}
    }
    \label{tab:performance_table}
\end{table}

\clearpage
\subsection{Ablation studies and systematic elucidation of failure modes}

To further elucidate the capabilities and limitations of \glspl{vllm} we created a set of tests intended to shed light on the strengths and limitations of these models. Most of these tests were created using the same images as for the main corpus of \macbench, but changing the textual part of the questions. \Cref{tab:ablations_description_table} describes each test, highlighting the differences from the original tasks.

\begin{table}
    \caption{\textbf{Descriptions for the different ablations performed.} Note that multi-step tasks are the same as some tasks in \macbench corpus. This is because multi-step reasoning is needed to solve the questions associated with these tasks.}
    \small
    \resizebox{\textwidth}{!}{
    \input{tables/ablation_table.tex}
    }
    \label{tab:ablations_description_table}
\end{table}

\paragraph{Performance}

\Cref{tab:ablations_performance_table} lists the performance in all our systematic failure mode elucidation experiments.

\begin{table}
    \caption{\textbf{Absolute performance of the different models in all the failure mode elucidation experiments.}}
    \resizebox{\textwidth}{!}{
    \input{tables/ablation_performance.tex}
    }
    \label{tab:ablations_performance_table}
\end{table}

\subparagraph{Crystal structure analysis}
\begin{figure}[htb]
    \centering
    \includegraphics[width=0.7\linewidth]{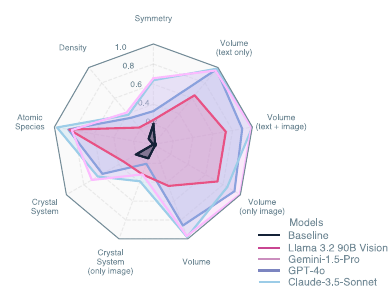}
    \caption{\textbf{\glspl{vllm} performance in tasks dealing with the interpretation of crystal structure renderings.} }
    \label{fig:cif_performance}
\end{figure}

In \Cref{fig:cif_performance} we show the performance of \glspl{vllm} on tasks concerning the analysis of crystal structures. To probe the influence of the modality on the performance, we showed the lattice parameters in only the text, only in the image, or in text and image. Interestingly, the performance changes depending on the modalities in which information is shown. 
In addition, the plot highlights that models show low performance for tasks requiring spatial reasoning, e.g., the assignment of the space group or crystal system. In the case of the assignment of the crystal system, we see that adding the lattice parameters to the image (which is by default included in all questions) helps the model perform better compared to only having access to the rendering of the structure.

\subparagraph{Organic chemistry performance}
\begin{figure}[htb]
    \centering
    \includegraphics[width=0.7\linewidth]{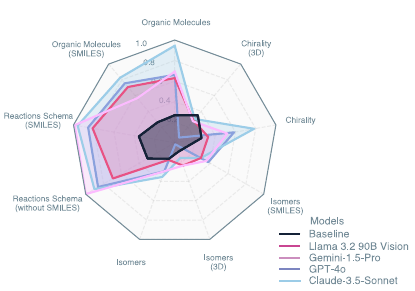}
    \caption{\textbf{\glspl{vllm} performance for questions related to organic molecules and reactions in \macbench.} }
    \label{fig:organic_performance}
\end{figure}

In \Cref{fig:organic_performance} we show the performance in tasks related to renderings of organic molecules and reactions. One of the most striking observations is the low performance in tasks related to identifying isomeric relationships between molecules. Here, the models perform comparably to the baseline in the vision modality and only slightly better than the baseline when provided with \gls{smiles} as text. 
We further observe that 3D-rendered molecular visualizations, generated using PyMOL \autocite{PyMOL}, result in reduced model performance compared to their 2D counterparts.
This trend is consistent across both Isomer and Chirality tasks, suggesting that spatial complexity in graphical representations may hinder model interpretation.

Similar limitations in spatial reasoning are probably the reason for low performance in tasks related to the assignment of chiral centers.

\subparagraph{Comparison with optical chemical structure recognition tools}

To establish a robust performance evaluation of \glspl{vllm} in chemical image analysis, we compared their effectiveness in the hand-drawn molecule recognition task (see \Cref{tab:domain_topics}) against Decimer\autocite{Rajan2021,Rajan2024}, a state-of-the-art tool designed explicitly for chemical structure recognition. 
This comparative analysis serves dual objectives: highlighting the relative strengths of general-purpose \glspl{vllm} against domain-specific tools while also assessing whether current \gls{vllm} capabilities meet the rigorous performance thresholds required of specialized systems in precision-critical scientific applications.

\begin{figure}[htb]
    \centering
    \includegraphics[width=\linewidth]{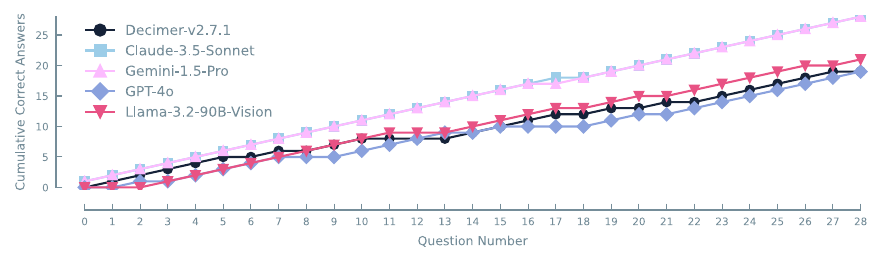}
    \caption{\textbf{Cumulative performance comparison between \glspl{vllm} and Decimer on hand-drawn molecular images.}}
    \label{fig:decimer_performance}
\end{figure}

As shown in \Cref{fig:decimer_performance}, the \glspl{vllm} (\claudethreefivesonnet and \geminipro) demonstrate superior performance compared to the specialized Decimer model in chemical structure recognition. 
This suggests that leading \glspl{vllm} can surpass technical models for specific cheminformatics tasks like molecule image interpretation.
Notably, the error analysis reveals consensus failures between top-performing \glspl{vllm} and Decimer.
These shared failure cases likely contain structurally complex molecules that present inherent challenges for current recognition systems, as evidenced by consistent performance drops across all models.
The correlation in error patterns implies that molecular complexity rather than model architecture limitations may be the primary factor in these challenging cases.

\clearpage
\subsection{Refusals}

By manually checking some of the answers for \macbench, we observed that some models refused to answer some of the questions, claiming they could not answer that type of question. This is probably a consequence of the safety alignment that the models go through.\autocite{cui2024orbenchoverrefusalbenchmarklarge} 
As a result of these observations, we counted the number of refusal response occurrences, which results are described in \Cref{tab:refusal_table}. Note that the results shown in the table include the original tasks and tests. Only the tasks for which some models refused are shown. Similarly, only the models that showed refusals are shown (\claudethreefivesonnet and \gptfouro).
Interestingly, we observe that \gptfouro refuses to answer many of the Lab Safety (50\%)  and CIF Structure Symmetry questions (49.5\%).

\begin{table}
    \caption{\textbf{Number of refused answers per topic for the \macbench corpus averaged over five different runs.} The percentages for each topic are relative to each task, while the overall percentage is relative to the total number of questions in the \macbench corpus. The topics in the table are the only ones for which refusal was observed. Similarly, note that the only models present are \claudethreefivesonnet, and \gptfouro because only these models showed refusals. For \gptfouro, a great percentage of the refusals are observed in Lab Safety and CIF Structure Symmetry questions, probably triggered because of the safety training of this model.}
    \resizebox{\textwidth}{!}{
    \input{tables/main_refusal.tex}
    }
    \label{tab:refusal_table}
\end{table}

\clearpage
\subsection{Sensitivity to prompt template} \label{sec:prompt_fragility}
\begin{figure}[htb]
    \centering
    \includegraphics[width=0.7\linewidth]{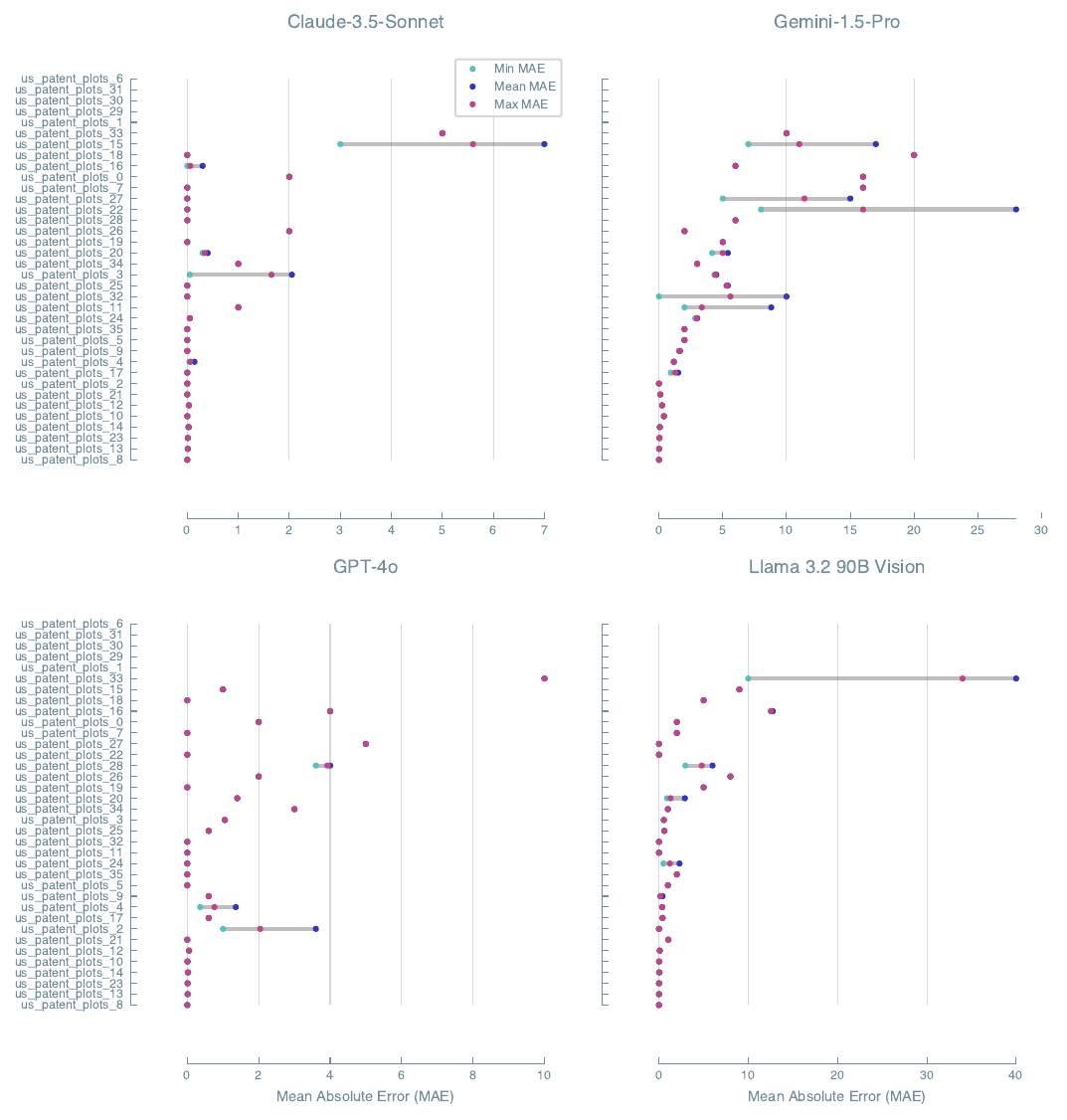}
    \caption{\textbf{\glspl{vllm} variation in performance with six different prompt templates for \uspplot questions in  \macbench.} 
    The rows list different questions from the \uspplot subset. The dumbbell plots show the statistics of the \gls{mae} for responses in different templates. The gray bars illustrate the variance; larger bars hence indicate larger sensitivity to the prompt template. Only questions that show variation for any model are included. Each question was run five times and averaged \gls{mae} was compared across templates.
    We observe that \geminipro is highly sensitive to the prompt template.}
    \label{fig:template_sensitivity}
\end{figure}

To study the sensitivity of \glspl{vllm} to prompt variations, we conducted a study in which we tested six template variations, differing only by a single word: \enquote{image} (original word), \enquote{diagram}, \enquote{plot}, \enquote{figure}, \enquote{photograph}, and \enquote{None} (leaving a space). 
In \Cref{fig:template_sensitivity} we show variation in \gls{mae} for the task considered for these tests.

\clearpage
\subsection{Sensitivity to system prompt}

To systematically evaluate the impact of prompt design variations in the \macbench benchmark, we developed five distinct system prompts representing different complexity levels, ranging from minimal instructional frameworks to comprehensive, context-rich formulations.\autocite{chen2024unleashingpotentialpromptengineering, chang2024efficientpromptingmethodslarge, xu2023compresspromptimprovingaccuracyefficiency, amatriain2024promptdesignengineeringintroduction}
This controlled experimental setup enables precise attribution of performance differences to specific prompt characteristics within the evaluation paradigm.\autocite{Han2024}

The system prompt serves to align the model with its intended operational environment.\autocite{zahan2025leveraginglargelanguagemodels}
A standard approach for crafting such prompts involves specifying a persona or role for the model and defining a task to contextualize its objective.\autocite{White2023, shen2023promptorconversationalautonomousprompt} 
The granularity of these specifications depends on both the complexity of the task and the model’s architecture. 
While longer, highly detailed prompts can enhance performance by clarifying expectations and constraints, they may also introduce inefficiencies due to the additional tokens that the model must process. 
This trade-off underscores the importance of balancing specificity and concision: overly brief prompts risk underspecifying the task, whereas excessively verbose prompts may dilute focus or introduce noise. 
Optimizing this equilibrium is critical to maximizing model efficacy.

Defining task specifications for a broad and diverse benchmark like \macbench presents inherent challenges due to significant variations in question types across its thematic categories. 
Excessively specific task descriptions risk introducing unintended biases, particularly if models overfit to granular details of individual tasks. 
For prompts 1–4, we adopt a generalized task definition framework, deliberately avoiding domain-specific instructions to minimize this bias.
To systematically evaluate how task-specific granularity influences benchmark performance metrics, we introduce prompt 5, which incorporates targeted definitions for a subset of \macbench tasks.\autocite{cao2024spider2vfarmultimodalagents}
This design enables direct comparison between general and fine-grained task formulations.

\paragraph{System prompt 1} \label{sec:prompt_1}
The initial prompt configuration (\Cref{lst:prompt_1}) establishes two key components: a domain-specific persona definition and a task characterization. 
For the persona specification, we assigned all five system prompt variations to PhD-level expertise in chemistry and materials science. 
This advanced academic profile was selected to ensure responses reflect specialized domain knowledge and critical thinking capabilities characteristic of doctoral researchers. 
The task was explicitly defined as a general question-answering (Q\&A) format, requiring models to synthesize technical information while maintaining accessibility for interdisciplinary scientific audiences.

\begin{lstlisting}[
  language=Python,
  caption={System prompt 1},
  label={lst:prompt_1}
]
You are a chemistry and materials expert with a PhD-level understanding, and your task is to answer some questions about those topics as accurately as possible.
\end{lstlisting}

\paragraph{System prompt 2} \label{sec:prompt_2}
The second prompt variation (\Cref{lst:prompt_2}) exclusively specifies the predefined persona without including explicit task instructions. This minimalist design isolates the effect of role specification, enabling two comparisons: against a baseline with no system prompt and against configurations combining both persona and task definitions. Thus, this ablation study framework systematically evaluates how doctoral-level domain expertise alone influences model outputs.

\begin{lstlisting}[
  language=Python,
  caption={System prompt 2},
  label={lst:prompt_2}
]
You are a chemistry and materials expert with a PhD-level understanding of those topics.
\end{lstlisting}

\paragraph{System prompt 3} \label{sec:prompt_3}
The third configuration (\Cref{lst:prompt_3}) retains the Q\&A task specification from Prompt 1 while removing all persona-related content. This complementary ablation isolates the task definition’s independent contribution by enabling comparisons against the persona-only configuration (\Cref{sec:prompt_2}) to quantify relative component impacts, and against the full persona-task configuration (\Cref{sec:prompt_1}) to assess whether domain expertise multiplicatively enhances task performance.

\begin{lstlisting}[
  language=Python,
  caption={System prompt 3},
  label={lst:prompt_3}
]
Your task is to answer some questions about chemistry and materials as accurately as possible.
\end{lstlisting}

\paragraph{System prompt 4} \label{sec:prompt_4}
The fourth configuration (\Cref{lst:prompt_4}) introduces an augmented task specification that explicitly addresses the multimodal nature of the benchmark. Unlike previous task definitions, this version contains explicit instruction for visual analysis of attached images and procedural guidance for integrating textual and visual information. By systematically varying only the task description while maintaining parity in other parameters, this design enables direct comparison with both the baseline task formulation (\Cref{sec:prompt_3}) and the combined persona-task configuration (\Cref{sec:prompt_1}), isolating the effects of granular task articulation on multimodal reasoning performance.

\newpage
\begin{lstlisting}[
  language=Python,
  caption={System prompt 4},
  label={lst:prompt_4}
]
You are a scientific assistant in a lab with a PhD-level understanding of chemical and material science principles.
You will be provided multiple-choice or exact-match questions to answer as accurately as possible.
The question will be related to chemistry and materials, and it will include an image that you must analyze to answer.
\end{lstlisting}

\paragraph{System prompt 5} \label{sec:prompt_5}
The final system prompt (\Cref{lst:prompt_5}) introduces detailed role specifications and explicit task guidelines through key components such as a comprehensive expert profile emphasizing multimodal analysis of chemical data, step-by-step guidelines for specific analytical tasks, or explicit implementation of advanced techniques like Chain-of-Thought reasoning\autocite{wei2023chainofthoughtpromptingelicitsreasoning} and self-verification mechanisms.\autocite{weng2023largelanguagemodelsbetter} The instruction incorporates task-specific focus areas mapped to analytical methods, such as pairing XRD pattern interpretation with crystal structure analysis, along with cross-modal validation requirements for text-image consistency checks and discipline-specific communication standards for technical reporting.

\begin{table}
    \caption{\textbf{Overall results for the different models with the different system prompts tested.}}
    \centering
    \input{tables/overall_system_table.tex}
    \label{tab:main_system_table}
\end{table}

\newpage
\begin{lstlisting}[
  language=Python,
  caption={System prompt 5},
  label={lst:prompt_5}
]
Act as an expert scientific assistant specializing in chemistry and materials science with advanced multimodal reasoning capabilities. You hold a PhD-level understanding of chemical principles, material characterization techniques, and analytical instrumentation, with particular expertise in:

Lab and experiments procedure: Laboratory safety and equipment operation.
Analytical Methods: Spectroscopy (FTIR, Raman, NMR), microscopy (SEM, TEM, AFM), diffraction techniques (XRD), and thermal analysis
Data Interpretation: Structure-property relationships, phase diagrams, reaction mechanisms, and materials performance evaluation

Your Task:

Multimodal Analysis: Rigorously examine provided images (e.g., chemical structures, spectra, micrographs, phase diagrams) alongside textual questions. Cross-reference visual patterns with chemical knowledge through:

Elemental/functional group identification in spectra.
Crystal structure analysis from diffraction patterns.
Morphology-property correlations in microscopy images.
Reaction pathway deduction from mechanism diagrams.

Reasoning Protocol:
a) Contextual Understanding: Identify key question components (concepts, required knowledge level, visual dependencies).
b) Evidence Integration: Combine image features with domain knowledge (e.g., match spectral peaks to reference databases).
c) Error Checking: Flag inconsistencies between visual data and question premises.
d) Confidence Calibration: Provide likelihood estimates when multiple interpretations exist.

Response Requirements:

For multiple-choice: Analyze all options systematically, eliminate distractors using first principles.
For exact-match: State answers with precision, referencing specific visual features.
Include brief rationale highlighting decisive evidence from both text and images.

Communication Standards:

Maintain precision: "The XRD pattern shows (hkl) planes consistent with BCC structure..."
Use visualization-specific language: "The broadening of peak X in the spectrum suggests..."
Prioritize chemical accuracy over linguistic elegance.

Error Prevention Mechanisms:

Cross-validate spectral assignments using multiple peak correlations.
Verify dimensional consistency in material property calculations.
Check for scale bar calibrations in microscopy images.
Confirm temporal/logical consistency in reaction sequences.
\end{lstlisting}

\subsubsection{System prompt ablation results}
To evaluate the impact of system prompts on model performance, the benchmark was executed a single time for each system prompt configuration using \claudethreefivesonnet, \geminipro, and \gptfouro. Notably, the API service for \llamathreetwoninety disallowed the inclusion of system prompts for multimodal questions, impeding its participation in this ablation study. The aggregate performance metrics for all models across system prompt variations are summarized in \Cref{tab:main_system_table}.

The results show that the relative ranking of models by performance remains consistent regardless of the system prompt employed. However, distinct trends emerge for individual models: the three evaluated models exhibit marginal performance gains when using system prompts, with the biggest improvement of 5\% by \claudethreefivesonnet and \gptfouro.

A topic-wise analysis reveals further nuances. For \claudethreefivesonnet (\Cref{tab:claude_system_table}), no single system prompt dominates across all topics. Intriguingly, omitting the system prompt yields superior results for certain topics, such as Organic Molecules and CIF Structure Volume. This pattern persists for \gptfouro (\Cref{tab:gpt_system_table}) and \geminipro (\Cref{tab:gemini_system_table}), where system prompts fail to produce consistent performance improvements. Across all models, the highest accuracy for specific topics frequently corresponds to evaluations conducted without system prompts, suggesting that task-specific context may outweigh the benefits of generalized prompting strategies.

\newpage

\begin{table}
    \caption{\textbf{Overall results for \claudethreefivesonnet with the different prompts.}}
    \centering
    \resizebox{\textwidth}{!}{
    \input{tables/claude-3-5-sonnet-20240620_system_table.tex}
    }
    \label{tab:claude_system_table}
\end{table}

\newpage

\begin{table}
    \caption{\textbf{Overall results for \geminipro with the different prompts.}}
    \centering
    \resizebox{\textwidth}{!}{
    \input{tables/gemini-1.5-pro_system_table.tex}
    }
    \label{tab:gemini_system_table}
\end{table}

\newpage

\begin{table}
    \caption{\textbf{Overall results for \gptfouro with the different prompts.}}
    \centering
        \resizebox{\textwidth}{!}{
    \input{tables/gpt-4o-2024-08-06_system_table.tex}
    }
    \label{tab:gpt_system_table}
\end{table}

\clearpage
\subsection{Leaderboard}

\begin{figure}[htb]
    \centering
    \includegraphics[width=\textwidth]{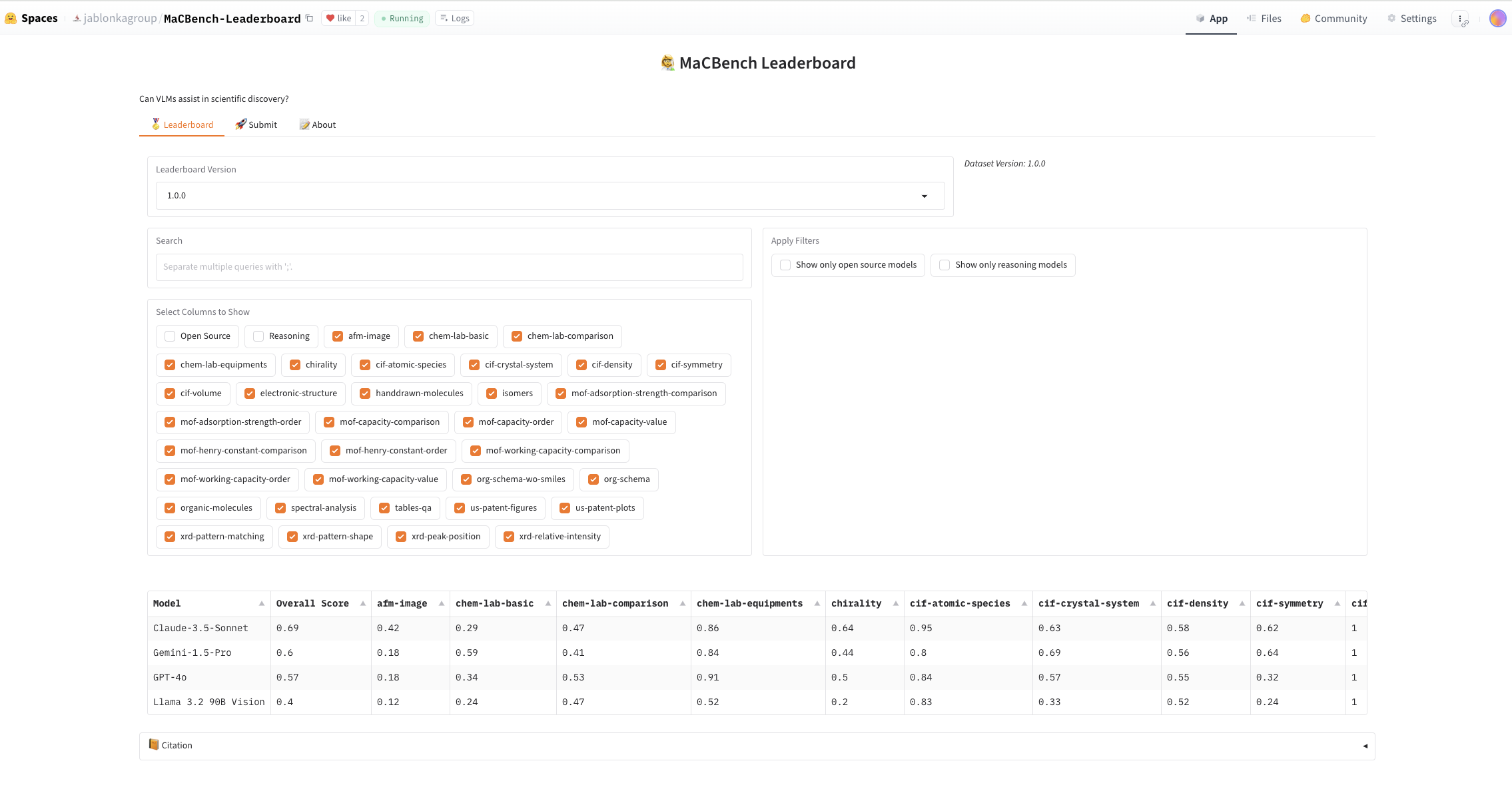}
    \caption{Screenshot of the leaderboard, which we deploy on HuggingFace.}
    \label{fig:leaderboard-screenshot}
\end{figure}

To summarize the results of \macbench, we created a leaderboard based on gradio and deployed it on HuggingFace Spaces (\Cref{fig:leaderboard-screenshot}). \autocite{abid2019gradio}
The online leaderboard is available at \url{https://huggingface.co/spaces/jablonkagroup/MaCBench-Leaderboard}.
We version the leaderboard using the HuggingFace dataset, which is version-controlled using git \autocite{lab_of_kevin_jablonka_at_uni_jena_2025_leaderboard}. Every time there is an update to the leaderboard, the version is bumped, and the user gets to select the version to view in the leaderboard UI.

\clearpage
\subsection{Recommendations for Improving VLLMs for Chemistry and Materials Science}

Based on the limitations identified in our \macbench evaluation, we propose several targeted recommendations to enhance the performance of multimodal models in scientific domains.

\subsubsection{Enhancing Spatial Reasoning Capabilities}

Our results revealed significant limitations in spatial reasoning tasks such as stereochemistry assignment and crystal structure interpretation. To address these limitations:

\begin{itemize}
    \item \textbf{Synthetic Training Data Generation:} Develop specialized datasets with explicit 3D spatial relationships labeled at various rotations and perspectives. For chemistry applications, this could include systematically generated stereoisomers and conformational variations.
    
    \item \textbf{Contrastive Learning Objectives:} Implement training objectives that require models to distinguish between subtle spatial differences (e.g., stereoisomers with identical 2D projections) to develop more robust spatial understanding.
\end{itemize}

\subsubsection{Improving Cross-Modal Integration}

The performance gap between text and image representations of identical information indicates weak cross-modal integration:

\begin{itemize}
    \item \textbf{Cross-Modal Alignment Training:} Design pretraining tasks requiring exact matching between chemical information represented in different modalities (e.g., spectral data in visual vs. tabular form).
    
    \item \textbf{Modality Translation Tasks:} Train models explicitly on tasks requiring transformation between representations (e.g., converting SMILES strings to structures and vice versa).
\end{itemize}

\subsubsection{Strengthening Multi-Step Reasoning}

The degradation in performance for tasks requiring multiple reasoning steps suggests needed improvements in this area:

\begin{itemize}
    \item \textbf{Test time inference:} Use chain-of-thought approaches to incentive the model to use more test-time compute.\autocite{snell2024scaling}
    
    \item \textbf{Reasoning Models:} Leverage recent advances in reasoning models or models with explicit reasoning components, which have shown promise in improving multi-step logical inference. These models specifically designed to strengthen step-by-step reasoning could help address the performance degradation we observed in tasks requiring chained analysis. This could involve fine-tuning on reasoning pathways\autocite{muennighoff2025s1} or reinforcement learning.\autocite{deepseek-ai2025deepseekr1} 

    \item \textbf{Tool-Augmented Architectures:} Implement modular architectures where specific components (such as external tools)\autocite{ramos2024review} handle different aspects of multi-step reasoning (e.g., one module for peak identification, another for relative ordering).

\end{itemize}

%% file: tables/description_table.tex
\begin{tabular}{lcp{12cm}}
\toprule
Topic & Description & Nº of Questions \\
\midrule
\midrule
\multicolumn{3}{l}{\textbf{Data Extraction}} \\
\midrule
Hand-drawn Molecules & 29 & Systematic naming of hand-drawn organic molecules \\
Organic Chemistry & & \\
\quad Chirality & 25 & Determination of the number of chiral centers in molecules, including their configuration, spatial orientation, and priority groups \\
\quad Isomers & 20 & Identification of isomeric relationships between two molecules \\
\quad Organic Molecules & 15 & Systematic naming of organic molecules following IUPAC nomenclature \\
\quad Organic Reaction Schema & 4 & Extraction of components such as solvents, temperature, or yield from organic reaction schemas \\
\quad Organic Reaction Schema without SMILES & 17 & Analysis of organic reaction schemas with visual references for molecule identification. \\
Tables and Plots & & \\
\quad Composition Tables & 308 & Analysis of composition tables \\
\quad US Patent Figures & 63 & Extraction of information from scientific figures in US patents \\
\quad US Patent Plots & 36 & Interpretation of 2D plots presented in US patents \\
\midrule
\multicolumn{3}{l}{\textbf{In silico and lab experiments}} \\
\midrule
Lab QA & & \\
\quad Lab Safety & 38 & Review of images taken in a chemistry lab focusing on safety protocols and proper laboratory practices \\
\quad Lab Safety Comparison & 17 & Comparison of laboratory images to identify correct practices and violations of good laboratory standards \\
Lab Equipments & 25 & Identification and classification of laboratory glassware and other equipment \\
CIF QA & & \\
\quad Crystal Structure Atomic Species & 41 & Determination of the number of different atomic species from crystal structure images \\
\quad Crystal Structure Density & 42 & Determination of the density from crystal structure images \\
\quad Crystal Structure Symmetry & 42 & Determination of the point group from crystal structure images \\
\quad Crystal Structure Volume & 42 & Determination of the volume from crystal structure images \\
\quad Crystal System & 42 & Determination of the crystal system from crystal structure images \\
\midrule
\multicolumn{3}{l}{\textbf{Data Interpretation}} \\
\midrule
AFM Image Analysis & 50 & Analysis of topography in various specimens using an atomic force microscope. \\
Adsorption Isotherm & & \\
\quad Adsorption Isotherm Capacity Comparison & 19 & Comparison of the capacities of adsorption isotherms \\
\quad Adsorption Isotherm Capacity Order & 20 & Ordering of capacities of adsorption isotherms \\
\quad Adsorption Isotherm Capacity Value & 20 & Determination of the capacity value from adsorption isotherms \\
\quad Adsorption Isotherm Henry Constant Comparison & 10 & Comparison of the Henry's constants of adsorption isotherms \\
\quad Adsorption Isotherm Henry Constant Order & 12 & Ordering of Henry's constants of adsorption isotherms \\
\quad Adsorption Isotherm Strength Comparison & 15 & Comparison of the adsorption strengths of isotherms \\
\quad Adsorption Isotherm Strength Order & 19 & Ordering of adsorption strengths of isotherms \\
\quad Adsorption Isotherm Working Capacity Comparison & 20 & Comparison of the working capacity of adsorption isotherms \\
\quad Adsorption Isotherm Working Capacity Order & 20 & Ordering of working capacities of adsorption isotherms \\
\quad Adsorption Isotherm Working Capacity Value & 20 & Determination of the working capacity value from adsorption isotherms \\
Electronic Structure & 24 & Analysis of the electronic structure of materials, such as direct or indirect bandgap and metallic characteristics \\
NMR and MS Spectra & 20 & Identification of halide atoms using MS isotope patterns and substitution positions on benzene rings using 1H NMR spectra \\
XRD QA & & \\
\quad XRD Pattern Matching & 20 & Determination of crystal type from a XRD pattern \\
\quad XRD Pattern Shape & 20 & Selection of the crystalline or amorphous nature from a XRD pattern \\
\quad XRD Peak Position & 20 & Determination of the peak position of most intense peak from a XRD pattern \\
\quad XRD Relative Intensity & 20 & Ordering of the peak positions of the three most intense peaks from XRD pattern \\
\midrule
Overall & 1155 & \\
\bottomrule
\end{tabular}

%% file: tables/main_results.tex
\begin{tabular}{lccccc}
\toprule
 & Baseline & Claude-3.5-Sonnet & Gemini-1.5-Pro & GPT-4o & Llama 3.2 90B Vision \\
\midrule
\midrule
\multicolumn{6}{l}{\textbf{Data Extraction}} \\
\midrule
Hand-drawn Molecules & 0.20±0.09 & \textbf{0.97±0.00} & \textbf{0.97±0.00} & 0.63±0.06 & 0.72±0.00 \\
Organic Chemistry & & & & & \\
\quad Chirality & 0.22±0.05 & \textbf{0.66±0.02} & 0.44±0.00 & 0.50±0.07 & 0.28±0.00 \\
\quad Isomers & 0.14±0.02 & \textbf{0.30±0.00} & 0.25±0.00 & 0.25±0.00 & 0.15±0.00 \\
\quad Organic Molecules & 0.23±0.11 & \textbf{0.80±0.00} & 0.59±0.03 & 0.56±0.04 & 0.53±0.00 \\
\quad Organic Reactions Schema & 0.15±0.22 & \textbf{1.00±0.00} & \textbf{1.00±0.00} & \textbf{1.00±0.00} & 0.50±0.00 \\
\quad Organic Reactions Schema without SMILES & 0.26±0.17 & 0.76±0.00 & \textbf{0.85±0.03} & 0.73±0.05 & 0.59±0.00 \\
Tables and Plots & & & & & \\
\quad Tables QA & 0.29±0.02 & \textbf{0.67±0.00} & 0.61±0.01 & 0.54±0.01 & 0.31±0.00 \\
\quad US Patent Figures & 0.15±0.03 & \textbf{0.67±0.00} & 0.32±0.01 & 0.54±0.01 & 0.27±0.00 \\
\quad US Patent Plots & 0.08±0.01 & \textbf{0.64±0.00} & 0.22±0.01 & 0.51±0.02 & 0.28±0.00 \\
\midrule
\multicolumn{6}{l}{\textbf{In Silico and Lab Experiments}} \\
\midrule
Lab QA & & & & & \\
\quad Lab Safety & 0.11±0.05 & 0.28±0.02 & \textbf{0.55±0.01} & 0.25±0.01 & 0.24±0.00 \\
\quad Lab Safety Comparison & 0.19±0.12 & \textbf{0.49±0.03} & 0.41±0.00 & 0.47±0.04 & 0.47±0.00 \\
Lab Equipments & 0.27±0.09 & 0.84±0.00 & 0.84±0.00 & \textbf{0.87±0.03} & 0.53±0.02 \\
CIF QA & & & & & \\
\quad CIF Structure Atomic Species & 0.00±0.00 & \textbf{0.95±0.00} & 0.81±0.01 & 0.82±0.03 & 0.83±0.00 \\
\quad CIF Structure Density & 0.07±0.00 & 0.39±0.04 & \textbf{0.40±0.05} & 0.31±0.06 & 0.21±0.00 \\
\quad CIF Structure Symmetry & 0.20±0.06 & 0.60±0.00 & \textbf{0.66±0.01} & 0.28±0.03 & 0.24±0.00 \\
\quad CIF Structure Volume & 0.02±0.00 & \textbf{0.96±0.01} & \textbf{0.96±0.02} & 0.83±0.02 & 0.43±0.00 \\
\quad CIF System & 0.20±0.07 & 0.53±0.02 & \textbf{0.69±0.00} & 0.57±0.00 & 0.33±0.00 \\
\midrule
\multicolumn{6}{l}{\textbf{Data Interpretation}} \\
\midrule
AFM Image Analysis & 0.00±0.00 & \textbf{0.43±0.01} & 0.21±0.02 & 0.21±0.03 & 0.14±0.00 \\
Adsorption Isotherm & & & & & \\
\quad Adsorption Isotherm Capacity Comparison & 0.31±0.11 & \textbf{0.99±0.02} & \textbf{0.99±0.02} & 0.88±0.02 & 0.63±0.00 \\
\quad Adsorption Isotherm Capacity Order & 0.24±0.11 & \textbf{0.85±0.00} & 0.76±0.04 & 0.63±0.03 & 0.55±0.00 \\
\quad Adsorption Isotherm Capacity Value & 0.27±0.10 & \textbf{0.74±0.02} & 0.65±0.00 & 0.44±0.05 & 0.55±0.00 \\
\quad Adsorption Isotherm Henry Constant Comparison & 0.22±0.08 & \textbf{1.00±0.00} & 0.64±0.09 & 0.88±0.04 & 0.80±0.00 \\
\quad Adsorption Isotherm Henry Constant Order & 0.27±0.12 & \textbf{0.82±0.04} & 0.67±0.00 & 0.75±0.00 & 0.75±0.00 \\
\quad Adsorption Isotherm Strength Comparison & 0.31±0.14 & \textbf{0.93±0.00} & 0.68±0.03 & 0.60±0.07 & 0.07±0.00 \\
\quad Adsorption Isotherm Strength Order & 0.35±0.06 & 0.74±0.00 & 0.49±0.03 & \textbf{0.78±0.02} & 0.37±0.00 \\
\quad Adsorption Isotherm Working Capacity Comparison & 0.36±0.10 & \textbf{0.76±0.04} & 0.55±0.00 & 0.53±0.04 & 0.45±0.00 \\
\quad Adsorption Isotherm Working Capacity Order & 0.23±0.06 & \textbf{0.71±0.07} & 0.50±0.00 & 0.64±0.02 & 0.55±0.00 \\
\quad Adsorption Isotherm Working Capacity Value & 0.20±0.09 & \textbf{0.67±0.03} & 0.33±0.08 & 0.24±0.11 & 0.25±0.00 \\
Electronic Structure & 0.46±0.10 & \textbf{0.70±0.02} & 0.39±0.00 & 0.56±0.02 & 0.39±0.00 \\
NMR and MS Spectra & 0.26±0.04 & 0.28±0.03 & 0.35±0.00 & \textbf{0.43±0.03} & 0.35±0.00 \\
XRD QA & & & & & \\
\quad XRD Pattern Matching & 0.25±0.08 & \textbf{0.52±0.03} & 0.27±0.03 & 0.28±0.04 & 0.45±0.00 \\
\quad XRD Pattern Shape & 0.31±0.08 & \textbf{0.89±0.02} & 0.71±0.02 & 0.85±0.00 & 0.30±0.00 \\
\quad XRD Peak Position & 0.23±0.10 & \textbf{1.00±0.00} & 0.85±0.00 & 0.80±0.04 & 0.30±0.00 \\
\quad XRD Relative Intensity & 0.23±0.11 & \textbf{0.46±0.06} & 0.35±0.03 & 0.17±0.02 & 0.16±0.00 \\
\midrule
\textbf{Overall} & 0.21±0.02 & \textbf{0.67±0.00} & 0.57±0.00 & 0.54±0.01 & 0.36±0.00 \\
\bottomrule
\end{tabular}

%% file: tables/ablation_table.tex
\begin{tabular}{lcp{12cm}}
\toprule
Ablation & N° of Questions & Description \\
\midrule
\midrule
\multicolumn{3}{l}{\textbf{Modality}} \\
\midrule
Crystal Structure Volume as Text & 42 & Calculation of crystal structure volume with lattice parameters given in text and image \\
Composition Tables (Ablation) & 308 & Evaluation of tabular data with text-based tuple representations instead of images. \\
XRD Pattern Matching  as Text & 20 & Determination of crystal type from a XRD pattern given as text \\
XRD Peak Position as Text & 20 & Determination of the peak position of most intense peak from a XRD pattern given the intensity and theta values as text \\
XRD Relative Intensity as Text & 20 & Ordering of the peak positions of the three most intense peaks from XRD pattern indicating in the text part of the question the intensity and theta values. \\
\midrule
\multicolumn{3}{l}{\textbf{Step}} \\
\midrule
Crystal Structure Density & 42 & Determination of the density from crystal structure images \\
Adsorption Isotherm Strength Order & 19 & Ordering of adsorption strengths of isotherms \\
Adsorption Isotherm Capacity Order & 20 & Ordering of capacities of adsorption isotherms \\
Adsorption Isotherm Henry Constant Order & 12 & Ordering of Henry's constants of adsorption isotherms \\
Adsorption Isotherm Working Capacity Order & 20 & Ordering of working capacities of adsorption isotherms \\
XRD Relative Intensity & 20 & Ordering of the peak positions of the three most intense peaks from XRD pattern \\
\midrule
\multicolumn{3}{l}{\textbf{Terminology}} \\
\midrule
Adsorption Isotherm Strength Comparison (Ablation) & 19 & Comparison of the adsorption strength of isotherms, avoiding scientific terminology \\
Adsorption Isotherm Strength Order (Ablation) & 18 & Ordering of adsorption strength of isotherms, avoiding scientific terminology \\
Adsorption Isotherm Capacity Comparison (Ablation) & 20 & Comparison of capacity of adsorption isotherms, avoiding scientific terminology \\
Adsorption Isotherm Capacity Order (Ablation) & 20 & Ordering of capacity of adsorption isotherms, avoiding scientific terminology \\
Adsorption Isotherm Capacity Value (Ablation) & 20 & Determination of the capacity value of adsorption isotherms, avoiding scientific terminology \\
Adsorption Isotherm Henry Constant Comparison (Ablation) & 10 & Comparison of Henry constants of adsorption isotherms, avoiding scientific terminology \\
Adsorption Isotherm Henry Constant Order (Ablation) & 10 & Ordering of the Henry constants of adsorption isotherms, avoiding scientific terminology \\
Adsorption Isotherm Working Capacity Comparison (Ablation) & 20 & Comparison of working capacities of adsorption isotherms, avoiding scientific terminology \\
Adsorption Isotherm Working Capacity Order (Ablation) & 20 & Ordering of the working capacity of adsorption isotherms, avoiding scientific terminology \\
Adsorption Isotherm Working Capacity Value (Ablation) & 20 & Determination of working capacity of adsorption isotherms, avoiding scientific terminology \\
US Patent Figures (Ablation) & 63 & Interpretation of patent figures avoiding the use of technical jargon. \\
US Patent Plots (Ablation) & 36 & Interpretation of patent plots with plain language, avoiding complex terminology. \\
XRD Pattern Shape (Ablation) & 20 & Adsorption isotherm pattern shape log (Ablation), avoiding scientific terminology \\
XRD Peak Position (Ablation) & 20 & Determination of the peak position in an XRD pattern with explanation on how to get this \\
XRD Relative Intensity (Ablation) & 20 & Ordering of the peak positions of the three most intense peaks from XRD pattern, avoiding scientific terminology \\
\midrule
\multicolumn{3}{l}{\textbf{Guidance}} \\
\midrule
Lab Safety (Guidance) & 38 & Examination of chemistry lab images with an emphasis on safety protocols, proper practices, and adherence to laboratory safety rules. \\
Electronic Structure with Knowledge & 24 & Investigation of electronic structures with instructions on how to solve the specific tasks \\
NMR and MS Spectra with Explanation & 20 & NMR and MS spectra with explanation on how to interpret these \\
XRD Pattern Matching (Ablation) & 20 & Determination of crystal type from a XRD pattern, avoiding scientific terminology \\
\midrule
\multicolumn{3}{l}{\textbf{Other}} \\
\midrule
AFM Image Analysis (Ablation) & 50 & Analysis of AFM images with additional details about legends, scales, and other image features \\
Chirality in 3D & 25 & Analysis of the chirality of a molecule in 3D \\
Crystal Structure System only image & 42 & Calculation of crystal structure volume without lattice parameters in image \\
Crystal Structure Volume without image & 42 & Calculate crystal structure volume with lattice parameters given in text without any image \\
Crystal Structure Volume parameters as image & 42 & Calculation of crystal structure volume with only lattice parameters in the image \\
Isomers in 3D & 15 & Study of isomeric relationships between two molecules in 3D \\
Isomers with SMILES & 20 & Analysis of isomeric relationships with SMILES representations for each molecule in the task description. \\
NMR and MS Spectra (Ablation) & 20 & Explicitly count the number of peaks or signals instead of elucidating the molecule of the spectra \\
Organic Molecules with SMILES & 10 & Systematic naming of organic molecules based on the SMILES \\
Organic Schema with SMILES & 16 & Analysis of organic reaction schemas using SMILES for molecule representation. \\
\midrule
Overall & 1263 & \\
\bottomrule
\end{tabular}

%% file: tables/ablation_performance.tex
\begin{tabular}{lccccc}
\toprule
 & Baseline & Claude-3.5 & Gemini-1.5 & GPT-4o & Llama-3.2 \\
\midrule
\midrule
\multicolumn{6}{l}{\textbf{Modality}} \\
\midrule
Crystal Structure Volume as Text & 0.02±0.00 & 0.95±0.00 & \textbf{0.99±0.01} & 0.90±0.01 & 0.74±0.00 \\
Composition Tables (Ablation) & 0.63±0.01 & \textbf{0.79±0.00} & 0.72±0.00 & 0.70±0.01 & 0.65±0.00 \\
XRD Pattern Matching  as Text & 0.35±0.06 & 0.59±0.04 & \textbf{0.65±0.04} & 0.44±0.04 & 0.54±0.02 \\
XRD Peak Position as Text & 0.26±0.07 & \textbf{1.00±0.00} & \textbf{1.00±0.00} & \textbf{1.00±0.00} & \textbf{1.00±0.00} \\
XRD Relative Intensity as Text & 0.25±0.09 & 0.39±0.02 & 0.35±0.00 & \textbf{0.44±0.02} & 0.28±0.04 \\
\midrule
\multicolumn{6}{l}{\textbf{Step}} \\
\midrule
Crystal Structure Density & 0.07±0.00 & 0.39±0.04 & \textbf{0.40±0.05} & 0.31±0.06 & 0.21±0.00 \\
Adsorption Isotherm Strength Order & 0.35±0.06 & 0.74±0.00 & 0.49±0.03 & \textbf{0.78±0.02} & 0.37±0.00 \\
Adsorption Isotherm Capacity Order & 0.24±0.11 & \textbf{0.85±0.00} & 0.76±0.04 & 0.63±0.03 & 0.55±0.00 \\
Adsorption Isotherm Henry Constant Order & 0.27±0.12 & \textbf{0.82±0.04} & 0.67±0.00 & 0.75±0.00 & 0.75±0.00 \\
Adsorption Isotherm Working Capacity Order & 0.23±0.06 & \textbf{0.71±0.07} & 0.50±0.00 & 0.64±0.02 & 0.55±0.00 \\
XRD Relative Intensity & 0.23±0.11 & \textbf{0.46±0.06} & 0.35±0.03 & 0.17±0.02 & 0.16±0.00 \\
\midrule
\multicolumn{6}{l}{\textbf{Terminology}} \\
\midrule
Adsorption Isotherm Strength Comparison (Ablation) & 0.22±0.08 & 0.84±0.00 & \textbf{0.86±0.03} & 0.53±0.04 & 0.21±0.00 \\
Adsorption Isotherm Strength Order (Ablation) & 0.40±0.10 & \textbf{1.00±0.00} & 0.76±0.03 & 0.88±0.07 & 0.83±0.00 \\
Adsorption Isotherm Capacity Comparison (Ablation) & 0.36±0.04 & 0.75±0.00 & \textbf{0.80±0.00} & 0.67±0.08 & 0.55±0.00 \\
Adsorption Isotherm Capacity Order (Ablation) & 0.26±0.09 & \textbf{0.95±0.00} & 0.85±0.00 & 0.90±0.04 & 0.60±0.00 \\
Adsorption Isotherm Capacity Value (Ablation) & 0.22±0.08 & \textbf{0.83±0.03} & 0.55±0.00 & 0.44±0.02 & 0.40±0.00 \\
Adsorption Isotherm Henry Constant Comparison (Ablation) & 0.24±0.09 & 0.89±0.00 & \textbf{1.00±0.00} & 0.89±0.00 & 0.56±0.00 \\
Adsorption Isotherm Henry Constant Order (Ablation) & 0.22±0.13 & \textbf{1.00±0.00} & \textbf{1.00±0.00} & 0.78±0.04 & 0.90±0.00 \\
Adsorption Isotherm Working Capacity Comparison (Ablation) & 0.25±0.09 & \textbf{0.89±0.02} & 0.78±0.03 & 0.76±0.02 & 0.80±0.00 \\
Adsorption Isotherm Working Capacity Order (Ablation) & 0.14±0.10 & 0.73±0.03 & 0.46±0.02 & \textbf{0.77±0.06} & 0.30±0.00 \\
Adsorption Isotherm Working Capacity Value (Ablation) & 0.22±0.09 & \textbf{0.95±0.00} & 0.46±0.04 & 0.51±0.05 & 0.45±0.00 \\
US Patent Figures (Ablation) & 0.11±0.04 & \textbf{0.63±0.02} & 0.33±0.01 & 0.51±0.02 & 0.27±0.01 \\
US Patent Plots (Ablation) & 0.06±0.03 & \textbf{0.59±0.02} & 0.11±0.00 & 0.39±0.00 & 0.28±0.00 \\
XRD Pattern Shape (Ablation) & 0.24±0.08 & \textbf{0.79±0.00} & 0.66±0.03 & \textbf{0.79±0.00} & 0.21±0.00 \\
XRD Peak Position (Ablation) & 0.34±0.08 & \textbf{0.91±0.02} & 0.90±0.00 & 0.87±0.03 & 0.75±0.00 \\
XRD Relative Intensity (Ablation) & 0.17±0.07 & \textbf{0.44±0.03} & 0.21±0.00 & 0.27±0.02 & 0.11±0.00 \\
\midrule
\multicolumn{6}{l}{\textbf{Guidance}} \\
\midrule
Lab Safety (Guidance) & 0.12±0.05 & 0.34±0.02 & \textbf{0.49±0.01} & 0.24±0.02 & 0.34±0.00 \\
Electronic Structure with Knowledge & 0.26±0.05 & 0.48±0.00 & 0.50±0.02 & \textbf{0.57±0.03} & 0.39±0.00 \\
NMR and MS Spectra with Explanation & 0.31±0.05 & 0.48±0.04 & 0.41±0.02 & \textbf{0.64±0.07} & 0.25±0.00 \\
XRD Pattern Matching (Ablation) & 0.31±0.10 & 0.42±0.04 & 0.45±0.00 & 0.36±0.05 & \textbf{0.50±0.00} \\
\midrule
\multicolumn{6}{l}{\textbf{Other}} \\
\midrule
AFM Image Analysis (Ablation) & 0.00±0.00 & \textbf{0.42±0.00} & 0.20±0.01 & 0.18±0.02 & 0.18±0.00 \\
NMR and MS Spectra (Ablation) & 0.25±0.07 & 0.50±0.00 & 0.68±0.04 & \textbf{0.75±0.04} & 0.55±0.00 \\
\midrule
\textbf{Overall} & 0.32±0.00 & \textbf{0.70±0.01} & 0.60±0.00 & 0.60±0.01 & 0.49±0.00 \\
\bottomrule
\end{tabular}

%% file: tables/main_refusal.tex
\begin{tabular}{lcccc}
\toprule
 & \multicolumn{2}{c}{Claude-3.5} & \multicolumn{2}{c}{GPT-4o} \\
\cmidrule(lr){2-3} \cmidrule(lr){4-5} \
 & Nº of Refusals & \% & Nº of Refusals & \% \\
\midrule
\midrule
\multicolumn{5}{l}{\textbf{Data Extraction}} \\
\midrule
Hand-drawn Molecules & 0.00±0.00 & 0.0 & 10.80±1.47 & 37.2 \\
Organic Chemistry & & & & \\
\quad Chirality & 2.40±0.49 & 9.6 & 1.80±0.75 & 7.2 \\
\quad Organic Molecules & 0.00±0.00 & 0.0 & 2.60±0.80 & 17.3 \\
Tables and Plots & & & & \\
\quad Tables QA & 0.00±0.00 & 0.0 & 0.60±0.49 & 0.2 \\
\quad US Patent Figures & 0.00±0.00 & 0.0 & 0.20±0.40 & 0.3 \\
\midrule
\multicolumn{5}{l}{\textbf{In Silico and Lab Experiments}} \\
\midrule
Lab QA & & & & \\
\quad Lab Safety & 0.00±0.00 & 0.0 & 19.00±1.79 & 50.0 \\
\quad Lab Safety Comparison & 0.00±0.00 & 0.0 & 3.80±1.47 & 22.4 \\
Lab Equipments & 0.00±0.00 & 0.0 & 1.20±0.75 & 4.8 \\
CIF QA & & & & \\
\quad CIF Structure Atomic Species & 0.00±0.00 & 0.0 & 2.00±1.10 & 4.9 \\
\quad CIF Structure Density & 12.60±0.80 & 30.0 & 0.20±0.40 & 0.5 \\
\quad CIF Structure Symmetry & 0.00±0.00 & 0.0 & 20.80±1.17 & 49.5 \\
\midrule
\multicolumn{5}{l}{\textbf{Data Interpretation}} \\
\midrule
AFM Image Analysis & 0.00±0.00 & 0.0 & 18.60±3.01 & 37.2 \\
Adsorption Isotherm & & & & \\
\quad Adsorption Isotherm Capacity Order & 0.00±0.00 & 0.0 & 0.20±0.40 & 1.0 \\
\quad Adsorption Isotherm Capacity Value & 0.00±0.00 & 0.0 & 5.40±1.96 & 27.0 \\
\quad Adsorption Isotherm Henry Constant Comparison & 0.00±0.00 & 0.0 & 0.20±0.40 & 2.0 \\
\quad Adsorption Isotherm Strength Comparison & 0.00±0.00 & 0.0 & 1.20±0.75 & 8.0 \\
\quad Adsorption Isotherm Strength Order & 0.00±0.00 & 0.0 & 0.20±0.40 & 1.1 \\
\quad Adsorption Isotherm Working Capacity Comparison & 0.00±0.00 & 0.0 & 0.20±0.40 & 1.0 \\
\quad Adsorption Isotherm Working Capacity Order & 0.00±0.00 & 0.0 & 0.40±0.49 & 2.0 \\
\quad Adsorption Isotherm Working Capacity Value & 0.00±0.00 & 0.0 & 1.20±1.47 & 6.0 \\
NMR and MS Spectra & 0.60±0.49 & 3.0 & 0.00±0.00 & 0.0 \\
XRD QA & & & & \\
\quad XRD Pattern Matching & 0.00±0.00 & 0.0 & 6.80±0.75 & 34.0 \\
\quad XRD Peak Position & 0.00±0.00 & 0.0 & 1.00±0.63 & 5.0 \\
\quad XRD Relative Intensity & 0.00±0.00 & 0.0 & 5.40±1.02 & 28.4 \\
\midrule
Overall & 15.60±1.78 & 1.4 & 103.80±22.25 & 9.0 \\
\bottomrule
\end{tabular}

%% file: tables/overall_system_table.tex
\begin{tabular}{lccc}
\toprule
Prompt & Claude-3.5-Sonnet & Gemini-1.5-Pro & GPT-4o \\
\midrule
\midrule
No system prompt & \textbf{0.67} & 0.57 & 0.54 \\
Prompt 1 & \textbf{0.68} & 0.58 & 0.54 \\
Prompt 2 & \textbf{0.69} & 0.6 & 0.56 \\
Prompt 3 & \textbf{0.67} & 0.58 & 0.54 \\
Prompt 4 & \textbf{0.69} & 0.58 & 0.57 \\
Prompt 5 & \textbf{0.71} & 0.56 & 0.61 \\
\bottomrule
\end{tabular}

%% file: tables/claude-3-5-sonnet-20240620_system_table.tex
\begin{tabular}{lcccccc}
\toprule
 & No Prompt & Prompt 1 & Prompt 2 & Prompt 3 & Prompt 4 & Prompt 5 \\
\midrule
\midrule
\multicolumn{7}{l}{\textbf{Data Extraction}} \\
\midrule
Hand-drawn Molecules & 0.97 & 0.97 & 0.97 & 0.97 & 0.97 & \textbf{1.00} \\
Organic Chemistry & & & & & & \\
\quad Chirality & \textbf{0.66} & 0.48 & 0.56 & 0.64 & 0.64 & 0.56 \\
\quad Isomers & 0.30 & \textbf{0.45} & 0.35 & 0.30 & 0.35 & 0.35 \\
\quad Organic Molecules & 0.80 & 0.80 & 0.80 & \textbf{0.87} & 0.80 & 0.80 \\
\quad Organic Reactions Schema & \textbf{1.00} & \textbf{1.00} & \textbf{1.00} & \textbf{1.00} & \textbf{1.00} & \textbf{1.00} \\
\quad Organic Reactions Schema without SMILES & 0.76 & 0.71 & \textbf{0.82} & 0.71 & 0.76 & \textbf{0.82} \\
Tables and Plots & & & & & & \\
\quad Tables QA & 0.67 & 0.71 & 0.70 & 0.70 & 0.72 & \textbf{0.75} \\
\quad US Patent Figures & \textbf{0.67} & 0.65 & \textbf{0.67} & \textbf{0.67} & \textbf{0.67} & 0.62 \\
\quad US Patent Plots & 0.64 & 0.67 & 0.69 & 0.67 & 0.72 & \textbf{0.75} \\
\midrule
\multicolumn{7}{l}{\textbf{In Silico and Lab Experiments}} \\
\midrule
Lab QA & & & & & & \\
\quad Lab Safety & 0.28 & 0.32 & \textbf{0.34} & 0.26 & \textbf{0.34} & 0.26 \\
\quad Lab Safety Comparison & \textbf{0.49} & 0.29 & 0.29 & 0.29 & 0.24 & 0.47 \\
Lab Equipments & 0.84 & 0.80 & 0.84 & 0.80 & \textbf{0.92} & 0.84 \\
CIF QA & & & & & & \\
\quad CIF Structure Atomic Species & \textbf{0.95} & 0.93 & \textbf{0.95} & 0.93 & \textbf{0.95} & \textbf{0.95} \\
\quad CIF Structure Density & 0.39 & 0.48 & 0.48 & 0.48 & 0.38 & \textbf{0.50} \\
\quad CIF Structure Symmetry & 0.60 & 0.62 & 0.57 & 0.50 & \textbf{0.69} & 0.67 \\
\quad CIF Structure Volume & \textbf{0.96} & 0.90 & 0.93 & 0.90 & 0.88 & 0.90 \\
\quad CIF System & 0.53 & 0.71 & 0.71 & 0.60 & 0.71 & \textbf{0.93} \\
\midrule
\multicolumn{7}{l}{\textbf{Data Interpretation}} \\
\midrule
AFM Image Analysis & 0.43 & 0.40 & 0.42 & 0.40 & 0.42 & \textbf{0.44} \\
Adsorption Isotherm & & & & & & \\
\quad Adsorption Isotherm Capacity Comparison & 0.99 & \textbf{1.00} & \textbf{1.00} & \textbf{1.00} & \textbf{1.00} & \textbf{1.00} \\
\quad Adsorption Isotherm Capacity Order & \textbf{0.85} & 0.80 & 0.80 & 0.80 & 0.80 & \textbf{0.85} \\
\quad Adsorption Isotherm Capacity Value & \textbf{0.74} & 0.70 & 0.65 & 0.70 & 0.70 & 0.70 \\
\quad Adsorption Isotherm Henry Constant Comparison & \textbf{1.00} & \textbf{1.00} & \textbf{1.00} & \textbf{1.00} & \textbf{1.00} & \textbf{1.00} \\
\quad Adsorption Isotherm Henry Constant Order & 0.82 & 0.83 & 0.83 & \textbf{0.92} & \textbf{0.92} & 0.83 \\
\quad Adsorption Isotherm Strength Comparison & \textbf{0.93} & \textbf{0.93} & \textbf{0.93} & \textbf{0.93} & \textbf{0.93} & \textbf{0.93} \\
\quad Adsorption Isotherm Strength Order & 0.74 & 0.74 & \textbf{0.79} & 0.74 & 0.74 & 0.74 \\
\quad Adsorption Isotherm Working Capacity Comparison & 0.76 & 0.75 & 0.80 & 0.75 & \textbf{0.90} & 0.80 \\
\quad Adsorption Isotherm Working Capacity Order & 0.71 & 0.70 & 0.80 & 0.75 & 0.65 & \textbf{0.90} \\
\quad Adsorption Isotherm Working Capacity Value & 0.67 & 0.55 & 0.70 & 0.55 & 0.55 & \textbf{0.80} \\
Electronic Structure & \textbf{0.70} & 0.61 & 0.61 & 0.65 & 0.57 & 0.39 \\
NMR and MS Spectra & 0.28 & 0.20 & 0.30 & 0.30 & 0.25 & \textbf{0.40} \\
XRD QA & & & & & & \\
\quad XRD Pattern Matching & 0.52 & \textbf{0.55} & 0.50 & 0.50 & 0.35 & 0.50 \\
\quad XRD Pattern Shape & 0.89 & 0.90 & 0.90 & 0.85 & \textbf{0.95} & 0.90 \\
\quad XRD Peak Position & \textbf{1.00} & \textbf{1.00} & \textbf{1.00} & \textbf{1.00} & \textbf{1.00} & 0.95 \\
\quad XRD Relative Intensity & 0.46 & \textbf{0.53} & \textbf{0.53} & 0.42 & \textbf{0.53} & 0.37 \\
\midrule
\textbf{Overall} & 0.67 & 0.68 & 0.69 & 0.67 & 0.69 & \textbf{0.71} \\
\bottomrule
\end{tabular}

%% file: tables/gemini-1.5-pro_system_table.tex
\begin{tabular}{lcccccc}
\toprule
 & No Prompt & Prompt 1 & Prompt 2 & Prompt 3 & Prompt 4 & Prompt 5 \\
\midrule
\midrule
\multicolumn{7}{l}{\textbf{Data Extraction}} \\
\midrule
Hand-drawn Molecules & 0.97 & \textbf{1.00} & 0.97 & 0.97 & 0.97 & \textbf{1.00} \\
Organic Chemistry & & & & & & \\
\quad Chirality & 0.44 & 0.44 & \textbf{0.52} & 0.44 & 0.48 & 0.44 \\
\quad Isomers & 0.25 & \textbf{0.30} & 0.25 & 0.25 & \textbf{0.30} & \textbf{0.30} \\
\quad Organic Molecules & 0.59 & 0.53 & \textbf{0.60} & \textbf{0.60} & 0.53 & 0.53 \\
\quad Organic Reactions Schema & \textbf{1.00} & \textbf{1.00} & \textbf{1.00} & \textbf{1.00} & \textbf{1.00} & \textbf{1.00} \\
\quad Organic Reactions Schema without SMILES & 0.85 & \textbf{0.88} & \textbf{0.88} & \textbf{0.88} & \textbf{0.88} & 0.76 \\
Tables and Plots & & & & & & \\
\quad Tables QA & 0.61 & 0.63 & \textbf{0.66} & 0.62 & 0.59 & 0.56 \\
\quad US Patent Figures & 0.32 & 0.29 & 0.32 & 0.30 & \textbf{0.35} & 0.33 \\
\quad US Patent Plots & \textbf{0.22} & 0.17 & 0.17 & 0.17 & 0.19 & 0.11 \\
\midrule
\multicolumn{7}{l}{\textbf{In Silico and Lab Experiments}} \\
\midrule
Lab QA & & & & & & \\
\quad Lab Safety & \textbf{0.55} & \textbf{0.55} & \textbf{0.55} & \textbf{0.55} & \textbf{0.55} & 0.47 \\
\quad Lab Safety Comparison & 0.41 & \textbf{0.47} & 0.41 & 0.41 & 0.41 & 0.41 \\
Lab Equipments & 0.84 & \textbf{0.88} & \textbf{0.88} & \textbf{0.88} & \textbf{0.88} & \textbf{0.88} \\
CIF QA & & & & & & \\
\quad CIF Structure Atomic Species & 0.81 & \textbf{0.83} & 0.80 & 0.80 & 0.80 & \textbf{0.83} \\
\quad CIF Structure Density & 0.40 & 0.33 & 0.36 & \textbf{0.43} & 0.38 & 0.24 \\
\quad CIF Structure Symmetry & 0.66 & 0.57 & 0.60 & 0.62 & 0.55 & \textbf{0.67} \\
\quad CIF Structure Volume & 0.96 & 0.95 & \textbf{0.98} & \textbf{0.98} & \textbf{0.98} & 0.95 \\
\quad CIF System & 0.69 & 0.76 & \textbf{0.88} & 0.64 & 0.64 & 0.62 \\
\midrule
\multicolumn{7}{l}{\textbf{Data Interpretation}} \\
\midrule
AFM Image Analysis & 0.21 & 0.16 & 0.20 & \textbf{0.22} & \textbf{0.22} & \textbf{0.22} \\
Adsorption Isotherm & & & & & & \\
\quad Adsorption Isotherm Capacity Comparison & \textbf{0.99} & 0.95 & 0.95 & 0.95 & 0.95 & 0.89 \\
\quad Adsorption Isotherm Capacity Order & 0.76 & 0.75 & 0.75 & 0.75 & 0.75 & \textbf{0.80} \\
\quad Adsorption Isotherm Capacity Value & 0.65 & 0.55 & 0.60 & 0.65 & 0.55 & \textbf{0.70} \\
\quad Adsorption Isotherm Henry Constant Comparison & 0.64 & 0.70 & \textbf{0.90} & 0.70 & 0.80 & 0.70 \\
\quad Adsorption Isotherm Henry Constant Order & 0.67 & \textbf{0.92} & 0.75 & 0.58 & 0.67 & 0.67 \\
\quad Adsorption Isotherm Strength Comparison & 0.68 & 0.80 & 0.80 & 0.73 & \textbf{0.87} & 0.60 \\
\quad Adsorption Isotherm Strength Order & 0.49 & 0.58 & 0.58 & 0.58 & \textbf{0.68} & 0.63 \\
\quad Adsorption Isotherm Working Capacity Comparison & 0.55 & 0.50 & 0.55 & 0.50 & 0.55 & \textbf{0.60} \\
\quad Adsorption Isotherm Working Capacity Order & 0.50 & 0.50 & 0.50 & \textbf{0.55} & 0.50 & 0.50 \\
\quad Adsorption Isotherm Working Capacity Value & 0.33 & 0.55 & 0.55 & \textbf{0.60} & 0.55 & 0.50 \\
Electronic Structure & 0.39 & 0.39 & 0.39 & 0.39 & 0.39 & \textbf{0.43} \\
NMR and MS Spectra & 0.35 & 0.35 & 0.35 & 0.35 & \textbf{0.40} & \textbf{0.40} \\
XRD QA & & & & & & \\
\quad XRD Pattern Matching & 0.27 & 0.25 & \textbf{0.35} & 0.25 & 0.30 & \textbf{0.35} \\
\quad XRD Pattern Shape & 0.71 & \textbf{0.75} & \textbf{0.75} & \textbf{0.75} & 0.70 & 0.70 \\
\quad XRD Peak Position & 0.85 & 0.85 & 0.85 & 0.85 & 0.85 & \textbf{0.90} \\
\quad XRD Relative Intensity & 0.35 & 0.26 & 0.26 & 0.26 & 0.26 & \textbf{0.47} \\
\midrule
\textbf{Overall} & 0.57 & 0.58 & \textbf{0.60} & 0.58 & 0.58 & 0.56 \\
\bottomrule
\end{tabular}

%% file: tables/gpt-4o-2024-08-06_system_table.tex
\begin{tabular}{lcccccc}
\toprule
 & No Prompt & Prompt 1 & Prompt 2 & Prompt 3 & Prompt 4 & Prompt 5 \\
\midrule
\midrule
\multicolumn{7}{l}{\textbf{Data Extraction}} \\
\midrule
Hand-drawn Molecules & 0.63 & 0.72 & 0.79 & 0.69 & \textbf{0.97} & 0.93 \\
Organic Chemistry & & & & & & \\
\quad Chirality & 0.50 & 0.52 & 0.56 & 0.52 & 0.56 & \textbf{0.68} \\
\quad Isomers & \textbf{0.25} & \textbf{0.25} & \textbf{0.25} & \textbf{0.25} & \textbf{0.25} & \textbf{0.25} \\
\quad Organic Molecules & 0.56 & 0.80 & \textbf{0.87} & 0.60 & 0.73 & 0.73 \\
\quad Organic Reactions Schema & \textbf{1.00} & \textbf{1.00} & \textbf{1.00} & \textbf{1.00} & \textbf{1.00} & \textbf{1.00} \\
\quad Organic Reactions Schema without SMILES & 0.73 & \textbf{0.76} & \textbf{0.76} & \textbf{0.76} & \textbf{0.76} & \textbf{0.76} \\
Tables and Plots & & & & & & \\
\quad Tables QA & 0.54 & 0.54 & 0.55 & 0.55 & 0.53 & \textbf{0.60} \\
\quad US Patent Figures & 0.54 & 0.54 & 0.49 & 0.52 & 0.54 & \textbf{0.56} \\
\quad US Patent Plots & \textbf{0.51} & 0.50 & 0.50 & 0.50 & 0.50 & 0.50 \\
\midrule
\multicolumn{7}{l}{\textbf{In Silico and Lab Experiments}} \\
\midrule
Lab QA & & & & & & \\
\quad Lab Safety & 0.25 & 0.29 & 0.39 & 0.26 & \textbf{0.45} & \textbf{0.45} \\
\quad Lab Safety Comparison & 0.47 & 0.47 & 0.53 & 0.53 & \textbf{0.59} & 0.47 \\
Lab Equipments & 0.87 & \textbf{0.92} & \textbf{0.92} & 0.88 & \textbf{0.92} & \textbf{0.92} \\
CIF QA & & & & & & \\
\quad CIF Structure Atomic Species & 0.82 & 0.71 & 0.76 & 0.78 & \textbf{0.88} & 0.83 \\
\quad CIF Structure Density & 0.31 & 0.33 & \textbf{0.40} & 0.33 & 0.31 & 0.33 \\
\quad CIF Structure Symmetry & 0.28 & 0.21 & 0.26 & 0.24 & 0.52 & \textbf{0.69} \\
\quad CIF Structure Volume & \textbf{0.83} & \textbf{0.83} & 0.79 & 0.81 & 0.76 & \textbf{0.83} \\
\quad CIF System & 0.57 & 0.60 & 0.62 & 0.57 & 0.57 & \textbf{0.88} \\
\midrule
\multicolumn{7}{l}{\textbf{Data Interpretation}} \\
\midrule
AFM Image Analysis & 0.21 & 0.18 & \textbf{0.22} & 0.18 & 0.18 & 0.16 \\
Adsorption Isotherm & & & & & & \\
\quad Adsorption Isotherm Capacity Comparison & 0.88 & \textbf{0.95} & 0.89 & 0.89 & \textbf{0.95} & \textbf{0.95} \\
\quad Adsorption Isotherm Capacity Order & 0.63 & 0.65 & 0.60 & 0.60 & 0.65 & \textbf{0.70} \\
\quad Adsorption Isotherm Capacity Value & 0.44 & 0.45 & 0.50 & 0.50 & \textbf{0.55} & 0.50 \\
\quad Adsorption Isotherm Henry Constant Comparison & 0.88 & \textbf{0.90} & \textbf{0.90} & \textbf{0.90} & \textbf{0.90} & \textbf{0.90} \\
\quad Adsorption Isotherm Henry Constant Order & 0.75 & 0.75 & \textbf{0.83} & \textbf{0.83} & 0.75 & 0.67 \\
\quad Adsorption Isotherm Strength Comparison & 0.60 & 0.60 & 0.60 & 0.60 & 0.60 & \textbf{0.67} \\
\quad Adsorption Isotherm Strength Order & 0.78 & \textbf{0.79} & \textbf{0.79} & \textbf{0.79} & \textbf{0.79} & 0.74 \\
\quad Adsorption Isotherm Working Capacity Comparison & 0.53 & 0.50 & 0.45 & 0.50 & 0.55 & \textbf{0.65} \\
\quad Adsorption Isotherm Working Capacity Order & 0.64 & 0.65 & \textbf{0.70} & 0.60 & 0.65 & 0.55 \\
\quad Adsorption Isotherm Working Capacity Value & 0.24 & 0.20 & 0.25 & 0.30 & \textbf{0.35} & 0.20 \\
Electronic Structure & 0.56 & \textbf{0.57} & \textbf{0.57} & \textbf{0.57} & 0.52 & 0.52 \\
NMR and MS Spectra & 0.43 & 0.45 & 0.40 & 0.40 & 0.40 & \textbf{0.50} \\
XRD QA & & & & & & \\
\quad XRD Pattern Matching & 0.28 & 0.25 & \textbf{0.35} & 0.30 & \textbf{0.35} & \textbf{0.35} \\
\quad XRD Pattern Shape & \textbf{0.85} & \textbf{0.85} & \textbf{0.85} & \textbf{0.85} & \textbf{0.85} & \textbf{0.85} \\
\quad XRD Peak Position & 0.80 & 0.80 & 0.80 & 0.80 & \textbf{0.85} & 0.80 \\
\quad XRD Relative Intensity & 0.17 & 0.21 & 0.21 & 0.21 & 0.32 & \textbf{0.53} \\
\midrule
\textbf{Overall} & 0.54 & 0.54 & 0.56 & 0.54 & 0.57 & \textbf{0.61} \\
\bottomrule
\end{tabular}